\documentclass[10pt,twocolumn,letterpaper]{article}

\usepackage[pagenumbers]{cvpr} 

\usepackage{graphicx}
\usepackage{amsmath}
\usepackage{amssymb}
\usepackage{booktabs}
\usepackage{multirow}
\usepackage{adjustbox}
\usepackage{soul}
\usepackage[table]{xcolor}
\usepackage{changepage,threeparttable}

\usepackage{mathtools}
\usepackage{bbding}
\usepackage{pifont}

\newcommand{\cmark}{\ding{51}}%
\newcommand{\xmark}{\ding{55}}%
\newcommand{\edit}[1]{#1}

\usepackage[pagebackref,breaklinks,colorlinks]{hyperref}

\usepackage[capitalize]{cleveref}
\crefname{section}{Sec.}{Secs.}
\Crefname{section}{Section}{Sections}
\Crefname{table}{Table}{Tables}
\crefname{table}{Tab.}{Tabs.}

\def\cvprPaperID{2248} 
\def\confName{CVPR}
\def\confYear{2024}

\begin{document}
	
	\title{MMSum: A Dataset for Multimodal Summarization \\
		and Thumbnail Generation of Videos}
	
	\author{Jielin Qiu$^{1,2}$, Jiacheng Zhu$^3$, William Han$^1$, Aditesh Kumar$^1$, Karthik Mittal$^1$, Claire Jin$^1$,\\
		Zhengyuan Yang$^2$, Linjie Li$^2$, Jianfeng Wang$^2$, Ding Zhao$^1$, Bo Li$^{4}$, Lijuan Wang$^2$\\
		$^1$Carnegie Mellon University, $^2$Microsoft Azure AI, $^3$MIT CSAIL, $^4$University of Chicago\\
		{\tt\footnotesize \{jielinq,wjhan,dingzhao\}@andrew.cmu.edu, zjc@mit.edu lbo@illinois.edu} \\ 
		{\tt\footnotesize \{zhengyang,lindsey.li,jianfw,lijuanw\}@microsoft.com}
	}
	\maketitle
	
	\begin{abstract}
		Multimodal summarization with multimodal output (MSMO) has emerged as a promising research direction. 
		Nonetheless, numerous limitations exist within existing public MSMO datasets, including insufficient maintenance, data inaccessibility, limited size, and the absence of proper categorization, which pose significant challenges.
		To address these challenges and provide a comprehensive dataset for this new direction, we have meticulously curated the \textbf{MMSum} dataset. 
		Our new dataset features 
		(1) Human-validated summaries for both video and textual content, providing superior human instruction and labels for multimodal learning.
		(2) Comprehensively and meticulously arranged categorization, spanning 17 principal categories and 170 subcategories to encapsulate a diverse array of real-world scenarios.
		(3) Benchmark tests performed on the proposed dataset to assess various tasks and methods, including \textit{video summarization}, \textit{text summarization}, and \textit{multimodal summarization}. 
		To champion accessibility and collaboration, we will release the \textbf{MMSum} dataset and the data collection tool as fully open-source resources, fostering transparency and accelerating future developments. 
		Our project website can be found at~\url{https://mmsum-dataset.github.io/}  
	\end{abstract}
	
	\section{Introduction}
	Multimodal summarization with multimodal output (MSMO) is an emerging research topic spurred by advancements in multimodal learning~\cite{Zhu2018MSMOMS,narasimhan2021clip,khullar2020mast,chen2018abstractive,hori2019end} and the increasing demand for real-world applications such as medical reporting~\cite{liu2021machine_mm_ehr}, educational materials~\cite{rahate2022multimodal_mm_edu}, and social behavior analysis~\cite{malhotra2020multimodal_mm_social}. 
	Most MSMO studies focus on video data and text data, aiming to select the most informative visual keyframes and condense the text content into key points.
	In this study, we focus on MSMO, which integrates both visual and textual information to provide users with comprehensive and representative summaries to enhance user experience~\cite{Zhu2018MSMOMS,li2020vmsmo,Fu2020MultimodalSF}. 
	
	\begin{figure}[tp]
		\centering
		\includegraphics[width=0.99\linewidth]{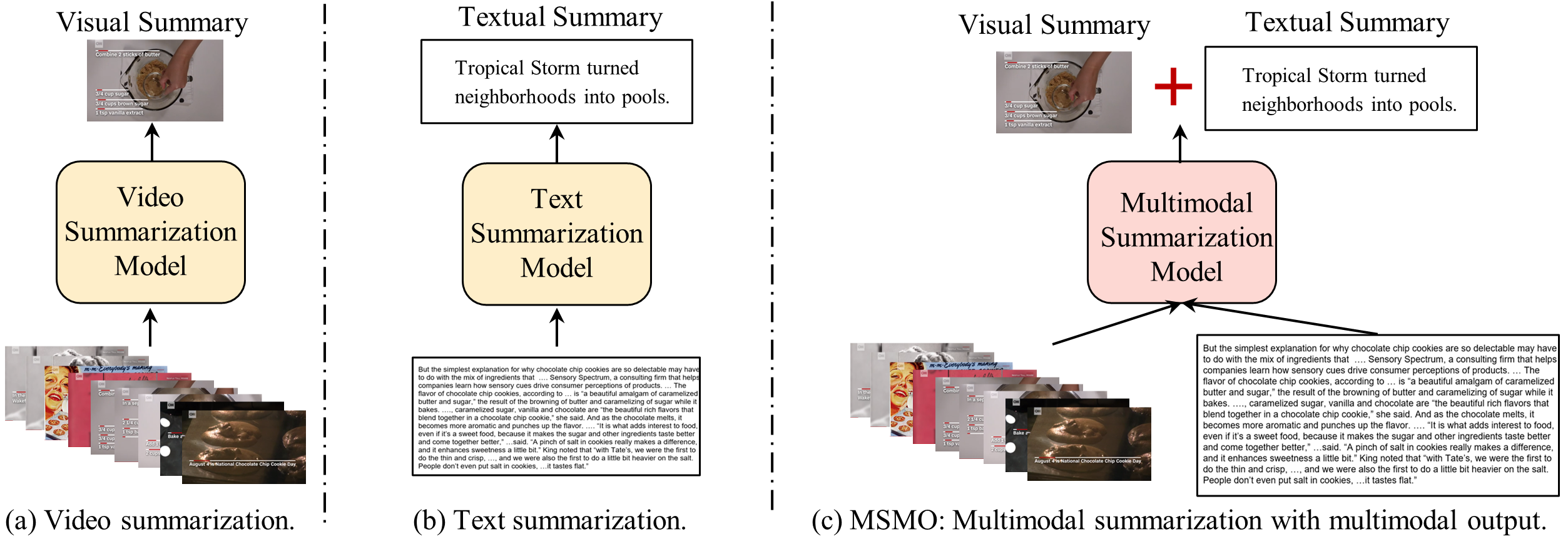}
		\caption{Task comparison of traditional video summarization, text summarization, and MSMO tasks.}
		\label{fig:task_compare}
		\vspace{-10pt}
	\end{figure}
	
	Despite the respective accomplishments of conventional unimodal summarization techniques on video data~\cite{zhang2016video,zhou2018deep,rochan2018video,zhu2020dsnet,park2020sumgraph,zhao2021reconstructive,jiang2022joint} and text data~\cite{nallapati2016abstractive,cheng2016neural,nallapati2017summarunner,miller2019leveraging,zhong2020extractive}, multimodal summarization continues to pose challenges due to a number of complexities. (1) The intricate nature of multimodal learning necessitates an algorithm capable of exploiting correlated information across different modalities, (2) There is a scarcity of appropriate multimodal datasets that reliably exhibit cross-modal correlations across diverse categories, and (3) There exists a gap in comprehensive evaluation protocols that accurately reflect the efficacy of MSMO methods in terms of their performance on both intermediate interpretations and downstream tasks.
	
	\begin{figure*}[tp]
		\centering
		\includegraphics[width=0.8\linewidth]{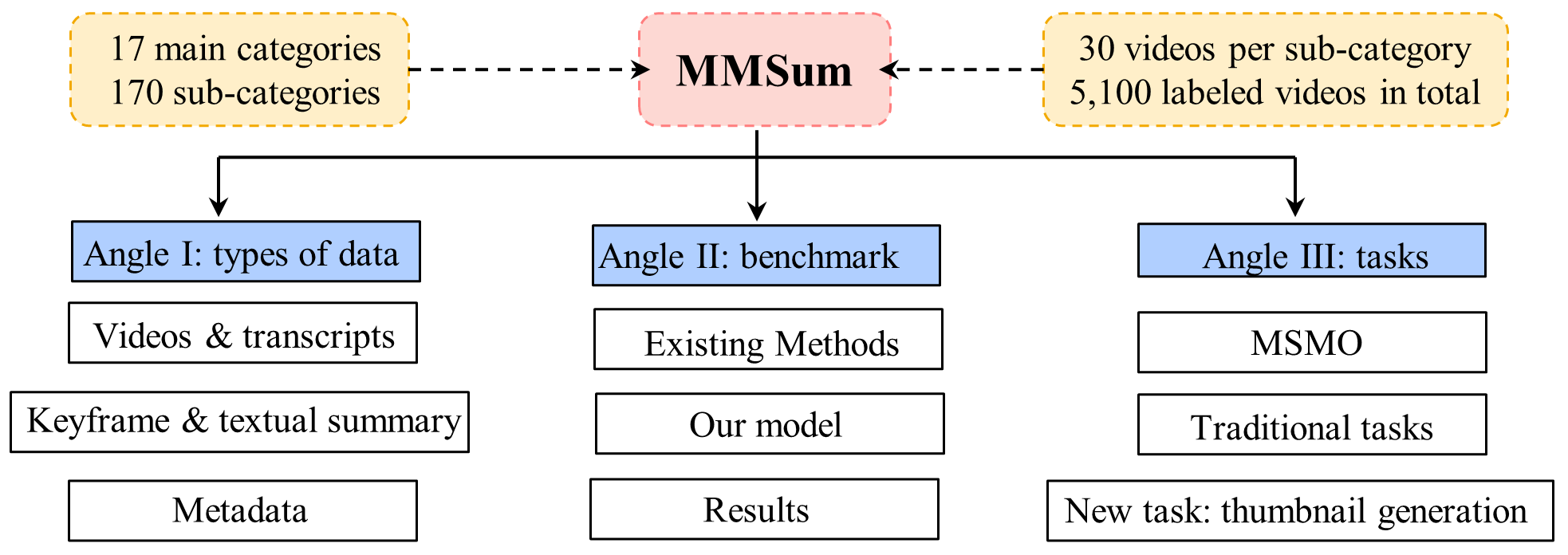}
		\caption{The design of the proposed MMSum dataset is driven by research and application needs.}
		\label{fig:pipeline}
		\vspace{-10pt}
	\end{figure*}
	
	Merging existing video and text datasets appears to be a feasible approach. However, assuring the presence of cross-modal correlations proves challenging~\cite{narasimhan2021clip}, not to mention the absence of necessary human verification~\cite{ouyang2022training_rlhf}, a vital element in machine learning research.
	Furthermore, the existing datasets pose several issues, such as inadequate maintenance leading to data unavailability, limited size, and lack of categorization. 
	To address these concerns and offer a comprehensive dataset for this area of study, we have undertaken the task of collecting a new dataset, named \textbf{MMSum}.
	Our contributions are summarised as follows:
	\vspace{-5pt}
	\begin{itemize}
		\item \textbf{A new MSMO dataset} Introducing MMSum, our newly curated MSMO dataset, specifically designed to cater to a wide range of tasks, with a particular emphasis on MSMO. This extensive dataset offers abundant information that serves as solid support for various research endeavors.
		\vspace{-5pt}
		\item \textbf{Diverse categorization} Within the MMSum dataset, we have meticulously gathered videos spanning 17 primary categories. Each of these main categories further comprises 10 distinct subcategories, culminating in a grand total of 170 subcategories. This comprehensive categorization ensures that the MMSum dataset is exceptionally representative and encompasses a wide range of content.
		\vspace{-5pt}
		\item \textbf{New benchmark} Across a diverse array of tasks, our results can be regarded as a benchmark on this novel real-world dataset.
		\vspace{-5pt}
		\item \textbf{Accessibility} We will open-source the MMSum dataset and the corresponding data collection tool with CC BY-NC-SA License.
	\end{itemize}

	\section{Related Work}
	
	\paragraph{Unimodal Summarization} typically comprises video summarization and text summarization. Video summarization involves extracting key moments that summarize the content of a video by selecting the most informative and essential parts. Traditional video summarization methods primarily rely on visual information. However, recent advancements have introduced category-driven or supervised approaches that generate video summaries by incorporating video-level labels, thereby enhancing the summarization process \cite{song2015tvsum,zhou2018deep,xiao2020convolutional,Zhou2018VideoSB,Haopeng2022ProgressiveVS,Narasimhan2022TLDWSI}. 
	Text Summarization involves processing textual metadata, such as documents, articles, tweets, and more, as input, and generating concise textual summaries. The quality of generated summaries has recently been significant improved through fine-tuning pre-trained language models \cite{liu-lapata-2019-text, zhang-etal-2019-hibert}.
	
	\vspace{-5pt}
	\paragraph{Multimodal Summarization} explored multiple modalities for summary generation. \cite{Otani2016VideoSU,Yuan2019VideoSB,Wei2018VideoSV,Fu2020MultimodalSF} learned the relevance or mapping in the latent space between different modalities.
	In addition to only generating visual summaries, \cite{Li2017MultimodalSF,Atri2021SeeHR,Zhu2018MSMOMS} generated textual summaries by taking audio, transcripts, or documents as input along with videos or images, using seq2seq model \cite{Sutskever2014SequenceTS} or attention mechanism \cite{Bahdanau2015NeuralMT}. The methods above explored using multiple modalities' information to generate single modality output, either textual or visual summary. 
	Recent trends on the MSMO task have also drawn much attention \cite{Zhu2018MSMOMS,Mingzhe2020VMSMOLT,Fu2021MMAVSAF,Fu2020MultimodalSF,Qiu2022MHMSMH,Zhang2022HierarchicalCS,Qiu2022SemanticsConsistentCS,Zhang2021UniMSAU,He2023AlignAA,Tang2022TLDWEM}.
	Specifically, \cite{Tang2022TLDWEM} summarized a video and text document into a cover frame and a one-sentence summary.
	The most significant difference between multimodal summarization and MSMO lies in the inclusion of multiple modalities in the output. 
	(More related work can be found in Appendix~\ref{sec:appendix-related-work}.)

	\begin{table*}[t]\small
		\centering
		\caption{Comparison of the modality of different \underline{summarization tasks and datasets}. Difference between traditional multimodal summarization and MSMO: traditional multimodal summarization still outputs a single-modality summary, while MSMO outputs both modalities' summaries. {Public Availability} means whether the data is still publicly available and valid. {Structural Summaries} means available summaries of each segment, not just for the whole video. 
		}
		\vspace{-5pt}
		\begin{adjustbox}{width=0.99\linewidth}
			\begin{tabular}{ l l|cc|cc |c|c|c}
\toprule
\multirow{2}{*}{Tasks} &\multirow{2}{*}{Datasets}  &\multicolumn{2}{c}{Input} &\multicolumn{2}{c}{Output} &\multirow{2}{*}{Public Availability} &\multirow{2}{*}{Categorization} &\multirow{2}{*}{Structural Summaries}\\
 & &\textbf{Visual} &\textbf{Textual} &\textbf{Visual} &\textbf{Textual}  &  & &  \\
\midrule
\multirow{3}{*}{Video} 
&TVSum \cite{Song2015TVSumSW}  & \cmark & \xmark & \cmark & \xmark & \cmark & \xmark & \cmark\\
&SumMe \cite{Gygli2014CreatingSF} & \cmark & \xmark & \cmark & \xmark & \cmark & \xmark & \cmark\\
& VSUMM \cite{de2011vsumm}  & \cmark & \xmark & \cmark & \xmark & \cmark & \xmark & \cmark\\
\cmidrule(l){1-1} \cmidrule(l){2-2} \cmidrule(l){3-4} \cmidrule(l){5-6} \cmidrule(l){7-7} \cmidrule(l){8-8} \cmidrule(l){9-9}
\multirow{2}{*}{Textual} 
&X-Sum \cite{Narayan2018DontGM} & \xmark & \cmark &\xmark  & \cmark & \cmark & \xmark & \xmark\\
&Pubmed \cite{Sen2008CollectiveCI} & \xmark & \cmark & \xmark  & \cmark & \cmark & \xmark & \xmark \\
\cmidrule(l){1-1} \cmidrule(l){2-2} \cmidrule(l){3-4} \cmidrule(l){5-6} \cmidrule(l){7-7} \cmidrule(l){8-8} \cmidrule(l){9-9}
\multirow{3}{*}{Multimodal} 
&How2 \cite{Sanabria2018How2AL}  & \cmark & \cmark & \cmark  & \xmark & \cmark & \xmark & \xmark\\
& AVIATE \cite{Atri2021SeeHR} & \cmark & \cmark & \xmark & \cmark & \cmark & \xmark & \xmark \\
& Daily Mail \cite{Zhu2018MSMOMS} & \cmark & \cmark & \xmark  & \xmark & \cmark & \xmark & \xmark \\
\cmidrule(l){1-1} \cmidrule(l){2-2} \cmidrule(l){3-4} \cmidrule(l){5-6} \cmidrule(l){7-7} \cmidrule(l){8-8} \cmidrule(l){9-9}
\multirow{3}{*}{MSMO} 
& VMSMO \cite{Mingzhe2020VMSMOLT} & \cmark & \cmark & \cmark & \cmark & \xmark & \xmark & \xmark \\
& MM-AVS \cite{Fu2020MultimodalSF} & \cmark & \cmark & \cmark & \cmark & \cmark & \xmark & \xmark \\
&\textbf{MMSum (Ours)}  & \textcolor{orange}{\cmark} & \textcolor{orange}{\cmark} & \textcolor{orange}{\cmark} & \textcolor{orange}{\cmark} & \textcolor{orange}{\cmark} & \textcolor{orange}{\cmark}  & \textcolor{orange}{\cmark} \\
\bottomrule
\end{tabular}
		\end{adjustbox}
		\label{Table:compare_tasks}
		\vspace{-10pt}
	\end{table*}

	\section{Angle I: Types of data}
	\subsection{Data Collection}

	In light of the aforementioned challenges inherent in the existing MSMO datasets, we propose a novel dataset named MMSum to address these issues comprehensively and effectively.
	Our approach involved the collection of a multimodal dataset, primarily sourced from a diverse range of untrimmed videos from YouTube. 
	The collected dataset comprises a rich set of information, including video files and transcripts, accompanied by corresponding video metadata. Additionally, temporal boundaries were meticulously recorded for each segment within the videos. Furthermore, for each segment, we obtained both video summaries and text summaries. It is worth noting that these summaries were directly provided by the authors of the respective videos, ensuring their authenticity and reliability.
	Moreover, the dataset incorporates comprehensive video metadata, such as titles, authors, URLs, categories, subcategories, and so on. 
	By gathering this diverse range of multimodal data and leveraging the ground-truth video and text summaries provided by the original content creators, we aim to create a valuable and reliable resource.

	\paragraph{Fidelity}

	Given the limited availability of fully annotated videos with complete and non-missing video summaries and text summaries, we resorted to a manual collection of videos that satisfied all the specified criteria. The meticulous nature of this process ensured that only videos meeting the stringent requirements were included in the dataset.
	To illustrate the disparities between different tasks and datasets in terms of modalities, we provide a comprehensive comparison in Table~\ref{Table:compare_tasks}. 
	For instance, traditional video or text summarization datasets typically encompass either visual or textual information exclusively. While there are datasets available for traditional multimodal summarization, where multiple modalities are used as input, they still produce single-modality summaries. In contrast, the MSMO dataset holds significant value in real-world applications, as it requires multimodal inputs and provides summaries containing both visual and textual elements. Consequently, the collection process for this dataset necessitates acquiring all the requisite information, resulting in a time-consuming endeavor.

	\paragraph{Human Verification}
	Notably, every video in the MMSum dataset undergoes manual verification to ensure high-quality data that fulfills all the specified requirements.
	For the fidelity verification process, five human experts (3 male and 2 female) each spent 30 days watching the collected videos, understanding the content, and verifying the annotations. The annotators were instructed to pay specific attention to the quality of segmentation boundaries, visual keyframes, and textual summaries. The pre-filtered size of the dataset is 6,800 (40 videos per subcategory). After manual verification and filtering, only 30 of 40 are preserved to ensure the quality, resulting in the current size of 5,100 (30 videos per subcategory).

	\begin{figure*}[t]
		\centering
		\begin{minipage}[t]{0.48\textwidth}
			\centering
			\includegraphics[width=0.99\linewidth]{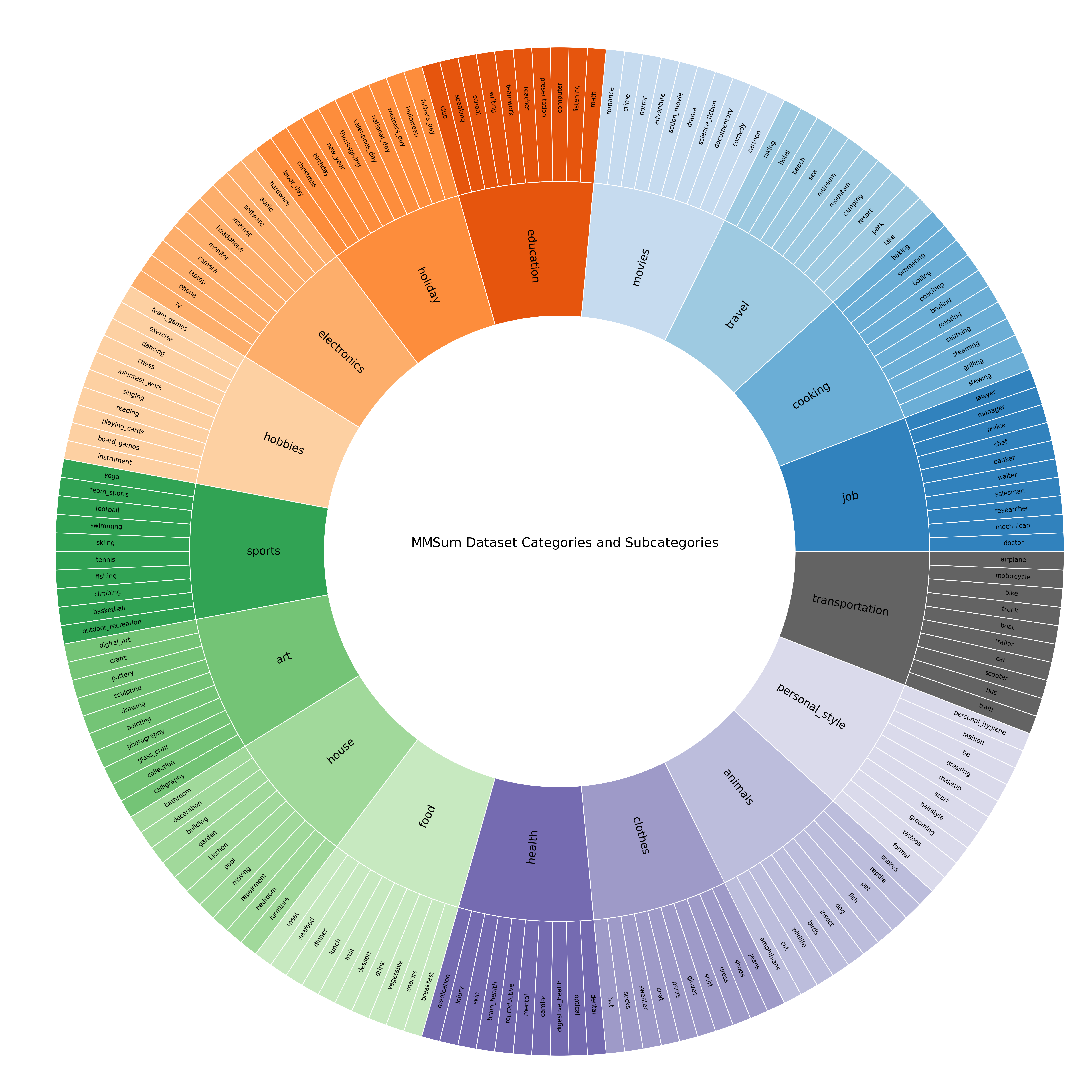}
			\caption{The 17 main categories of the MMSum dataset, where each main category contains 10 subcategories, resulting in 170 subcategories in total. More details are listed in Table~\ref{Table:category}. }
			\label{fig:categories}
		\end{minipage}
		~~
		\begin{minipage}[t]{0.48\textwidth}
			\centering
			\includegraphics[width=0.99\linewidth]{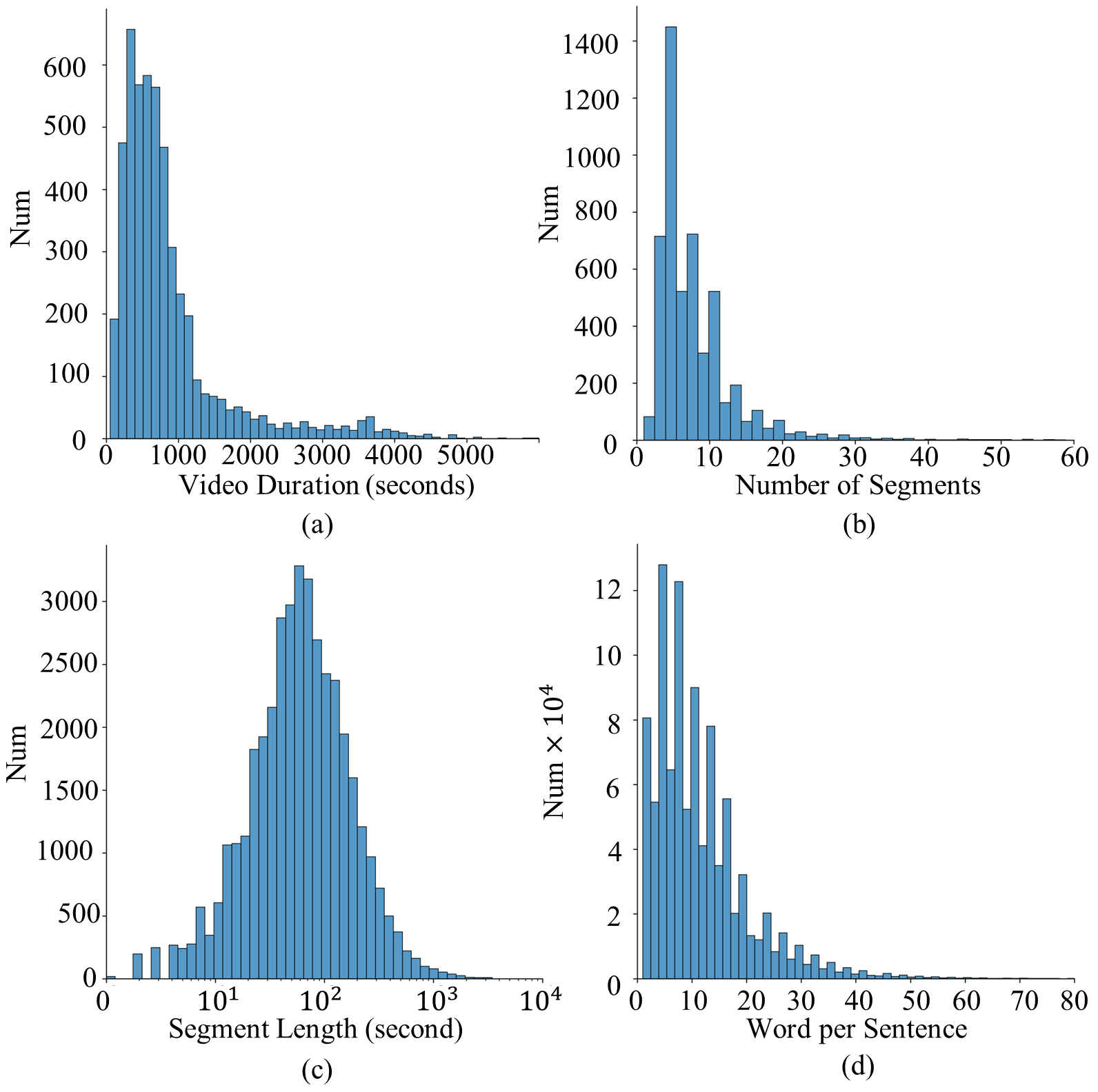}
			\caption{The statistics of the MMSum dataset, which show the distribution of (a) video duration; (b) number of segments
				per video; (c) segment duration; (d) number of words per sentence.}
			\label{fig:statistics}
		\end{minipage}
	\end{figure*}

	\begin{table*}[t]
		\caption{Comparison with existing video summarization and multimodal summarization datasets.}
		\vspace{-5pt}
		\centering
		\begin{adjustbox}{width=0.99\linewidth}
			\begin{tabular}{lcccccc}
				\toprule
				& SumMe \cite{Gygli2014CreatingSF} & TVSum \cite{Song2015TVSumSW} & OVP \cite{Avila2011VSUMMAM} & CNN \cite{Fu2021MMAVSAF} & Daily Mail \cite{Fu2021MMAVSAF}  & Ours  \\
				\midrule
				Source & YouTube & YouTube & YouTube & News & News & YouTube\\ 
				Number of Data &  25 & 50 &50 & 203 & 1,970 & \textbf{5,100}   \\
				Total Video Duration (Hours) & 1.0 & 3.5 &1.3 & 7.1 & 44.2 & \textbf{1229.9}  \\
				Average Video Duration (mins) &2.4 & 3.9 &1.6 &2.1 &1.4 & \textbf{14.5}\\
				Max Video Duration (mins)  &5.4 &10.8 &3.5 &6.9 &4.8 & \textbf{115.4}\\
				Min Video Duration (mins)  &0.5 &1.4 &0.8 &0.3 &0.4 & 1.0 \\
				Total Number of Text Tokens & -- & -- & -- & 0.2M & 1.3M & \textbf{11.2M} \\
				Avg. Keyframes per video & 44 & 70 & 9.6 & 7.1 & 2.9 & 7.8 \\
				Avg. Text Summary Length & -- & -- & -- & 29.7 & 59.6 & 21.69 \\   
				Number of Classes &25 &10 &7 &-- &-- & \textbf{170} \\
				\bottomrule
			\end{tabular}
		\end{adjustbox}
		\label{Table:compare_datasets}
		\vspace{-10pt}
	\end{table*}

	\vspace{-5pt}
	\paragraph{Diversity}

	During the dataset creation process, we extensively examined existing video datasets such as \cite{miech19howto100m,ZhXuCoAAAI18} for reference. Subsequently, we carefully selected \underline{17} main categories to ensure comprehensive coverage of diverse topics. These main categories encompass a wide range of subjects, including \textit{animals, education, health, travel, movies, cooking, job, electronics, art, personal style, clothes, sports, house, food, holiday, transportation}, and \textit{hobbies}.
	Each main category is further divided into \underline{10} subcategories based on the popularity of Wikipedia, resulting in a total of \underline{170} subcategories. To illustrate the subcategories associated with each main category, please refer to Figure~\ref{fig:categories} and Table~\ref{Table:category} (in the Appendix). For a more detailed view, a high-resolution version of Figure~\ref{fig:categories} can be found in Appendix~\ref{sec:appendix_categories}.
	To ensure the dataset's representativeness and practicality, we imposed certain criteria for video inclusion. Specifically, we only collected videos that were longer than 1 minute in duration while also ensuring that the maximum video duration did not exceed 120 minutes. Adhering to these guidelines allows a balance between capturing sufficient content in each video and preventing excessively lengthy videos from dominating the dataset.
	In total, our dataset comprises 170 subcategories and a grand total of 5,100 videos, all carefully selected to encompass a wide range of topics and characteristics.



	\subsection{Statistics of the Dataset}

	Figure~\ref{fig:statistics} presents a comprehensive analysis of the MMSum dataset's statistics. Figure~\ref{fig:statistics}(a) delves into the distribution of video durations, revealing the average duration spans approximately 15 minutes. In Figure~\ref{fig:statistics}(b), we show the distribution of the number of segments per video. The graph in Figure~\ref{fig:statistics}(c) captures the distribution of segment durations, showcasing an intriguing resemblance to the Gaussian distribution with an approximate mean of 80 seconds. Figure~\ref{fig:statistics}(d) shows the distribution of the number of words per sentence.

	\subsection{Comparison with Existing Datasets}
	
	Table~\ref{Table:compare_datasets} presents a comparison between our MMSum dataset and existing video datasets. In contrast to standard video summarization datasets such as SumMe \cite{Gygli2014CreatingSF}, TVSum \cite{Song2015TVSumSW}, and OVP \cite{Avila2011VSUMMAM}, our dataset, MMSum, stands out in several aspects. Firstly, the existing datasets lack textual data, whereas MMSum incorporates both video and textual information. Additionally, while the number of videos in SumMe, TVSum, and OVP is under 50, MMSum contains a substantial collection of 5,100 videos. Furthermore, the average duration of the videos in the aforementioned datasets is less than 4 minutes, whereas the videos in MMSum have an average duration of 14.5 minutes. Moreover, MMSum provides a significantly larger number of segments/keyframes per video compared to these standard datasets, making it more suitable for real-world applications.
	Comparing MMSum with other MSMO datasets like CNN and Daily Mail \cite{Fu2021MMAVSAF}, we find that our dataset first surpasses them in terms of the number of videos. Furthermore, CNN and Daily Mail datasets were not curated based on specific classes; instead, the data was randomly downloaded, resulting in a lack of representativeness. In contrast, MMSum was carefully designed with 17 main categories and 170 subcategories, making it highly representative and practical.
	Although there are other MSMO datasets like VMSMO \cite{Mingzhe2020VMSMOLT}, we did not include them in the comparison table due to a large portion of the video links no longer be valid. Therefore, MMSum stands out as a comprehensive and reliable dataset for multimodal summarization tasks. The key distinguishing features of MMSum can be summarized as follows:
	\vspace{-5pt}
	\begin{itemize}
		\item MMSum offers an extensive and large-scale dataset, comprising an impressive collection of 5,100 human-annotated videos.
		\vspace{-5pt}
		\item The dataset showcases a remarkable range of untrimmed videos, varying in duration from concise 1-minute clips to extensive recordings spanning up to 115 minutes. This diversity allows for a comprehensive exploration of different video lengths and content complexities.
		\vspace{-5pt}
		\item MMSum's strength lies in its meticulously crafted main category and subcategory groups, which exhibit an exceptional level of richness and granularity. With a keen focus on real-world applicability, these categories are thoughtfully designed to encapsulate the diverse facets and contexts of video data, ensuring relevance across a wide array of domains.
		\vspace{-5pt}
		\item To guarantee the highest quality and integrity of the dataset, MMSum undergoes rigorous manual verification. This meticulous process ensures that all modalities and information within the dataset are accurately annotated and readily accessible.
	\end{itemize}
	
	\begin{figure*}[t]
		\centering
		\includegraphics[width=0.99\linewidth]{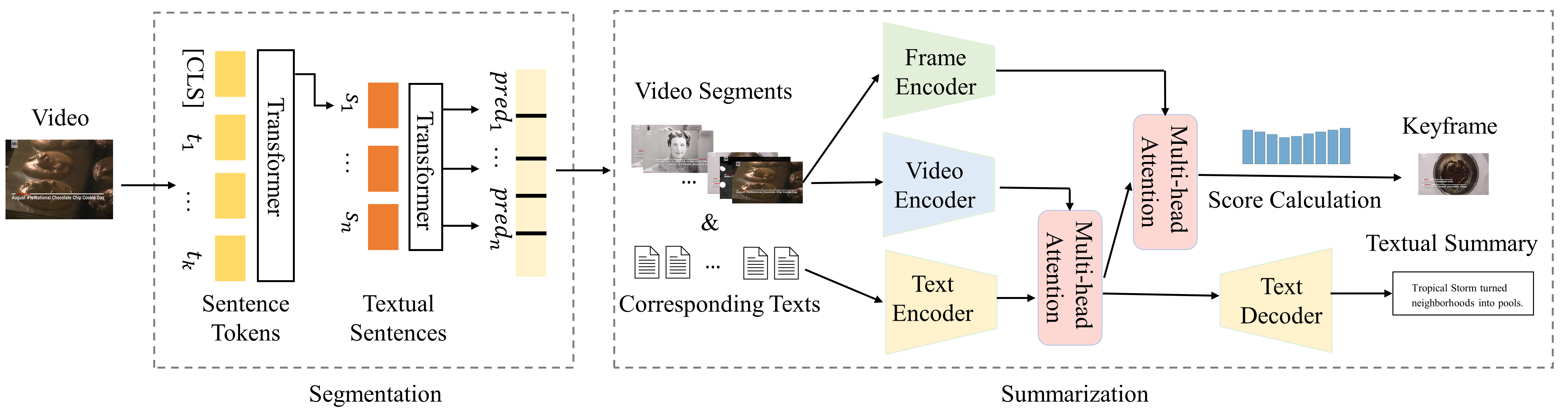}
		\vspace{-5pt}
		\caption{Our model comprises two modules: the segmentation module and the summarization module.}
		\label{fig:model}
		\vspace{-10pt}
	\end{figure*}

	\section{Angle II: Benchmark }
	
	\subsection{Problem Formulation}
	
	The formulation of the MSMO task can be expressed as follows. A video and its corresponding transcripts are denoted as a pair $(V, X)$. The video input, represented by $V$, consists of a sequence of frames: $V = (v_1, v_2, \ldots, v_N)$. The corresponding transcripts, denoted as $X$, are a sequence of sentences: $X = (x_1, x_2, \ldots, x_M)$,. Note that $M$ may not equal $N$ due to one sentence per frame is not guaranteed in real-world videos. It is assumed that each video has a sequence of ground-truth textual summary, denoted as $Y = (y_1, y_2, \ldots, y_L)$, and a sequence of ground-truth keyframe  represented by $P = (p_1, p_2, \ldots, p_L)$, where $L$ is the number of segments. The objective of the MSMO task is to generate textual summaries $\widehat{Y}$ that capture the main points of the video, and select keyframes $\widehat{P}$ from $V$ to be the visual summaries.
	
	\subsection{Existing Methods}
	In order to conduct a thorough performance evaluation, we selected a set of established methods as our baselines. These baselines are chosen based on the public availability of official implementations, ensuring reliable and reproducible results. 
	The selected baseline methods encompass:
	\vspace{-5pt}
	\begin{itemize}
		\vspace{-2pt}
		\item For Video Summarization: Uniform Sampling \cite{jadon2019video}, K-means Clustering \cite{Hartigan1979AKC}, VSUMM \cite{de2011vsumm}, and Keyframe Extraction \cite{jadon2019video}.
		\vspace{-10pt}
		\item For Text Summarization: BERT2BERT \cite{turc2019}, BART \cite{Lewis2020BARTDS} (BART-large-CNN and BART-large-XSUM), Distilbart \cite{Shleifer2020PretrainedSD}, T5 \cite{Raffel2019ExploringTL}, Pegasus  \cite{Zhang2019PEGASUSPW}, and LED \cite{Beltagy2020LongformerTL}.
	\end{itemize}
	\vspace{-5pt}
	More details of the baselines within the benchmark can be found in Appendix~\ref{sec:appendix-baselines}. However, due to the absence of publicly available implementations for MSMO methods in the existing literature, there are no suitable methods that can be used as MSMO baselines.

	\subsection{Our Method}

	To solve the problem mentioned above and provide a MSMO baseline for the collected MMSum dataset, we propose a novel and practical approach to augment the MSMO baseline. Our method, which we have made accessible on our website, comprises two modules: segmentation and summarization. Our model is depicted in Figure~\ref{fig:model}.
	
	\vspace{-5pt}
	\paragraph{Segmentation Module}
	
	The primary objective of the segmentation module is to partition a given video into smaller segments based on the underlying content. This module operates by leveraging the entire transcript associated with the video, employing a contextual understanding of the text.
	For the segmentation module, we adopted a hierarchical BERT architecture, which has demonstrated state-of-the-art performance \cite{lukasik-etal-2020-text}. It comprises two transformer encoders. The first encoder focuses on sentence-level encoding, while the second encoder handles paragraph-level encoding.
	The first encoder encodes each sentence independently using $\text{BERT}_{\text{LARGE}}$ and then feeds the encoded embeddings into the second encoder. Notably, all sequences commence with a special token [CLS] to facilitate encoding at the sentence level. If a segmentation decision is made at the sentence level, the [CLS] token is utilized as input for the second encoder, which enables inter-sentence relationships to be captured through cross-attention mechanisms. This enables a cohesive representation of the entire transcript, taking into account the contextual dependencies between sentences.

	\vspace{-5pt}
	\paragraph{Summarization Module}
	
	Upon segmenting the video, each video segment becomes the input to the summarization module. In line with the model architecture proposed in \cite{Krubiski2023MLASKMS}, we construct our summarization module.
	The summarization module incorporates three main encoders: a frame encoder, a video encoder, and a text encoder. These encoders are responsible for processing the video frames, video content, and corresponding text, respectively, to extract relevant feature representations.
	Once the features have been extracted, multi-head attention is employed to fuse the learned features from the different encoders, which allows for the integration of information across the modalities, enabling a holistic understanding of the video and its textual content.
	Following the fusion of features, a score calculation step is performed to select the keyframe, identifying the most salient frame within each video segment. Additionally, a text decoder is utilized to generate the textual summary, leveraging the extracted features and the fused representations.
	Considering our primary focus on providing a benchmark in this work, we have included model details in Appendix~\ref{sec:appendix-model} due to page limit.

	\section{Angle III: Tasks and Results }
	
	\subsection{Types of tasks}
	
	Within our dataset, a wealth of information is available, enabling the exploration of various downstream tasks. These tasks encompass video summarization (VS), text summarization (TS), and multimodal video summarization with multimodal output (MSMO).
	To provide a comprehensive understanding of each task and highlight their distinctions, we have compiled detailed descriptions and comparisons in Appendix~\ref{sec:tasks}. 
	For the train/val/test split, since our dataset is already randomly collected from YouTube, we designate the last 30\% of videos within each subcategory (indexed 21-29) as the testing set. The remaining videos are then assigned to the training set (indexed 00-20) in each subcategory. More results are shown in Appendix~\ref{sec:appendix-more-results}.

	\subsection{Evaluation of Traditional Tasks }

	\paragraph{Video Summarization Evaluation} The quality of the chosen keyframe is evaluated by Root Mean Squared Error (RMSE), Structural Similarity Index (SSIM), Signal reconstruction error ratio (SRE), and Spectral angle mapper (SAM), between image references and the extracted video frames \cite{Mller2020SuperresolutionOM}. In addition, we also adopted precision, recall, and F1 score based on SSIM for evaluation. 
	\vspace{-10pt}
	\paragraph{Text Summarization Evaluation} The quality of generated textual summary is evaluated by standard evaluation metrics, including BLEU~\cite{Papineni2002BleuAM}, METEOR~\cite{Denkowski2014MeteorUL}, ROUGE-L~\cite{Lin2004ROUGEAP}, CIDEr~\cite{Vedantam2015CIDErCI}, and BertScore \cite{Zhang2019BERTScoreET}, following previous works \cite{Abigail2017,Chen2018IterativeDR,Mingzhe2020VMSMOLT}. ROUGE-1, ROUGE-2, and ROUGE-L refer to the overlap of unigram, bigrams, and the longest common subsequence between the decoded summary and the reference, respectively \cite{Lin2004ROUGEAP}. 
	
	\begin{table*}[t]\small
		\caption{Comparison of video summarization results (whole-video setting and segment-level setting).}
		\vspace{-5pt}
		\centering
		\begin{adjustbox}{width=0.9\linewidth}
			\begin{tabular}{llcccc|ccc}
\toprule
Setting &Model  & RMSE $\downarrow$  & PSNR $\uparrow$ &SSIM $\uparrow$    &SRE $\downarrow$   &Precision $\uparrow$ &Recall $\uparrow$ & F1 Score $\uparrow$ \\ 
\midrule
\multirow{4}{*}{Whole-video}
&Uniform \cite{jadon2019video} & 0.479 & 4.044 & 0.076 & 49.808 & 0.077 & 0.100 & 0.049\\
&K-means \cite{Hartigan1979AKC} &0.348 &8.234 &0.055 &46.438 &0.072 &0.182 &0.103  \\   
&VSUMM \cite{de2011vsumm} &0.279 &9.226 &0.053 &44.862 &0.054 &0.259 &0.088 \\
&Ours&\textbf{0.112}&\textbf{25.280}&\textbf{0.697}&\textbf{23.550}&\textbf{0.320}&\textbf{0.290} & \textbf{0.321}\\
\midrule
\multirow{4}{*}{Segment-level}
&Uniform \cite{jadon2019video} & 0.237 & 6.307 & 0.085 & 42.495 & 0.186 & 0.179 & 0.105\\
&K-means \cite{Hartigan1979AKC} &0.167 &10.123 &0.144 &46.533 &0.123 &0.172 &0.143   \\
&VSUMM \cite{de2011vsumm} & 0.122 & 18.818 & 0.258 & 41.601 & 0.160 & 0.207 & 0.171\\
&Ours &\textbf{0.091}&\textbf{36.370}&\textbf{0.698}&\textbf{23.430}&\textbf{0.333}&\textbf{0.275}&\textbf{0.255}\\
\bottomrule
\end{tabular}

		\end{adjustbox}
		\label{table:vs_results}
		\vspace{-5pt}
	\end{table*}

	\begin{table*}[t]\small
		\caption{Comparison of textual summarization results (whole-video setting and segment-level setting).}
		\vspace{-2pt}
		\centering
		\begin{adjustbox}{width=0.99\linewidth}
			\begin{tabular}{llrrrrrrrr}
\toprule
Setting &Model  & BLEU-1 $\uparrow$ & ROUGE-1 $\uparrow$ & ROUGE-2 $\uparrow$  & ROUGE-L $\uparrow$ & METEOR $\uparrow$ & CIDEr $\uparrow$ & SPICE $\uparrow$ &BertScore $\uparrow$ \\
\midrule
\multirow{8}{*}{Whole-video}
&BERT2BERT \cite{turc2019} & 22.59  & 3.75 & 0.45 & 3.41 & 5.65 & 1.76 & 2.91 & 71.12 \\
&BART-large-CNN \cite{Lewis2020BARTDS} & 29.17  & 3.19 & 0.51 & 3.04 & 2.99 & 2.28 & 11.27 & 68.84\\
&BART-large-XSUM \cite{Lewis2020BARTDS} & 30.91  & 3.83 & 0.57 & 3.59 & 3.99 & 2.56 & 3.71 & 69.56 \\
&Distilbart \cite{Shleifer2020PretrainedSD}  & 26.46  & 3.87 & 3.87 & 0.47 & 3.59 & 2.25 & 4.16 & 69.37 \\
&T5 \cite{Raffel2019ExploringTL} & 25.39  & 3.51 & 0.43 & 3.21 & 4.51 & 1.97 & 5.66 & 70.38\\
&Pegasus \cite{Zhang2019PEGASUSPW} & 26.73  & 3.75 & 0.52 & 3.40 & 4.52 & 2.38 & 7.82 & 68.92 \\
&LED \cite{Beltagy2020LongformerTL} & 26.47  & 3.81 & 0.25 & 3.51 & 3.45 & 1.78 & 6.72 & 68.45\\
&Ours &  \textbf{32.61} & \textbf{9.41} & \textbf{2.86} & \textbf{9.15} & 4.01 & 4.01 & 10.11 & \textbf{74.46}\\
\midrule
\multirow{8}{*}{Segment-level}
&BERT2BERT \cite{turc2019} & 13.58  & 4.70 & 1.95 & 4.53 & 28.59 & 11.73 & 10.13 & 71.76 \\
&BART-large-CNN \cite{Lewis2020BARTDS} & 22.79  & 6.45 & 2.46 & 6.32 & 26.21 & 20.64 & 10.13 & 71.44\\ 
&BART-large-XSUM \cite{Lewis2020BARTDS} & 20.89  & 7.31 & 2.77 & 7.13 & 29.36 & 20.90 & 10.20 & 71.42 \\
&Distilbart \cite{Shleifer2020PretrainedSD} & 14.77  & 1.95 & 0.15 & 1.87 & 23.52 & 11.83 & 10.53 & 66.46\\
&T5 \cite{Raffel2019ExploringTL} & 16.48  & 6.17 & 3.03 & 5.99 & 28.22 & 20.96 & 10.35 & 71.95 \\
&Pegasus \cite{Zhang2019PEGASUSPW} & 16.17  & 3.41 & 0.96 & 3.29 & 29.82 & 17.26 & 10.39 & 67.81\\
&LED \cite{Beltagy2020LongformerTL} & 16.03  & 3.80 & 0.60 & 3.64 & 29.81 & 15.85 & 10.99 & 68.46\\
&Ours &  \textbf{23.36} & \textbf{13.61} & \textbf{4.58} & \textbf{13.24} & \textbf{30.01} & \textbf{21.06} & 10.28 & \textbf{85.19}\\ 
\bottomrule
\end{tabular}

			\label{table:ts_results}
		\end{adjustbox}
		\vspace{-10pt}
	\end{table*}
	
	\begin{table*}[t]\small
		\caption{Comparison of MSMO results.}
		\vspace{-5pt}
		\centering
		\begin{adjustbox}{width=0.99\linewidth}
			\begin{tabular}{l|cccccccccccc} 
\toprule
\multirow{2}{*}{Methods}   & \multicolumn{5}{c}{Text} &\multicolumn{5}{c}{Video}   \\ 
\cmidrule(r){2-6}\cmidrule(r){7-11} 
& BLEU $\uparrow$ & METEOR $\uparrow$ & CIDEr $\uparrow$ & SPICE $\uparrow$ &BertScore $\uparrow$ &PSNR $\uparrow$ &SSIM $\uparrow$ &Precision  $\uparrow$ &Recall $\uparrow$ &F1 Score $\uparrow$ \\ 
\midrule
LGSS + VSUMM + T5 & 27.35 & 24.32 & 3.94 & 5.57 & 62.77 &16.234  &0.198  &0.143  &0.152  &0.147   \\ 
LGSS +  VSUMM + BART-large-XSUM & 24.83 & 24.12 & 3.97 & 8.86  &39.20 &16.234  &0.198  &0.143  &0.152  &0.147    \\
LGSS +  VSUMM + BERT2BERT & 13.26 & 24.83 & 3.68 & 9.23& 64.34  &16.234  &0.198  &0.143  &0.152  &0.147   \\
LGSS +  VSUMM + BART-large-CNN & 24.93 & 28.61 & 3.78 & 9.84 & 64.44 &16.234  &0.198  &0.143  &0.152  &0.147 \\
Ours& \textbf{33.36} & \textbf{30.31} & \textbf{4.06} & \textbf{10.28} & \textbf{85.19} & \textbf{36.370} & \textbf{0.298} & \textbf{0.233} & \textbf{0.275} &  \textbf{0.155} \\
\bottomrule
\end{tabular}
			\label{table:msmo_results}
		\end{adjustbox}
	\end{table*}
	
	\begin{figure*}[htp]
		\centering
		\includegraphics[width=0.88\linewidth]{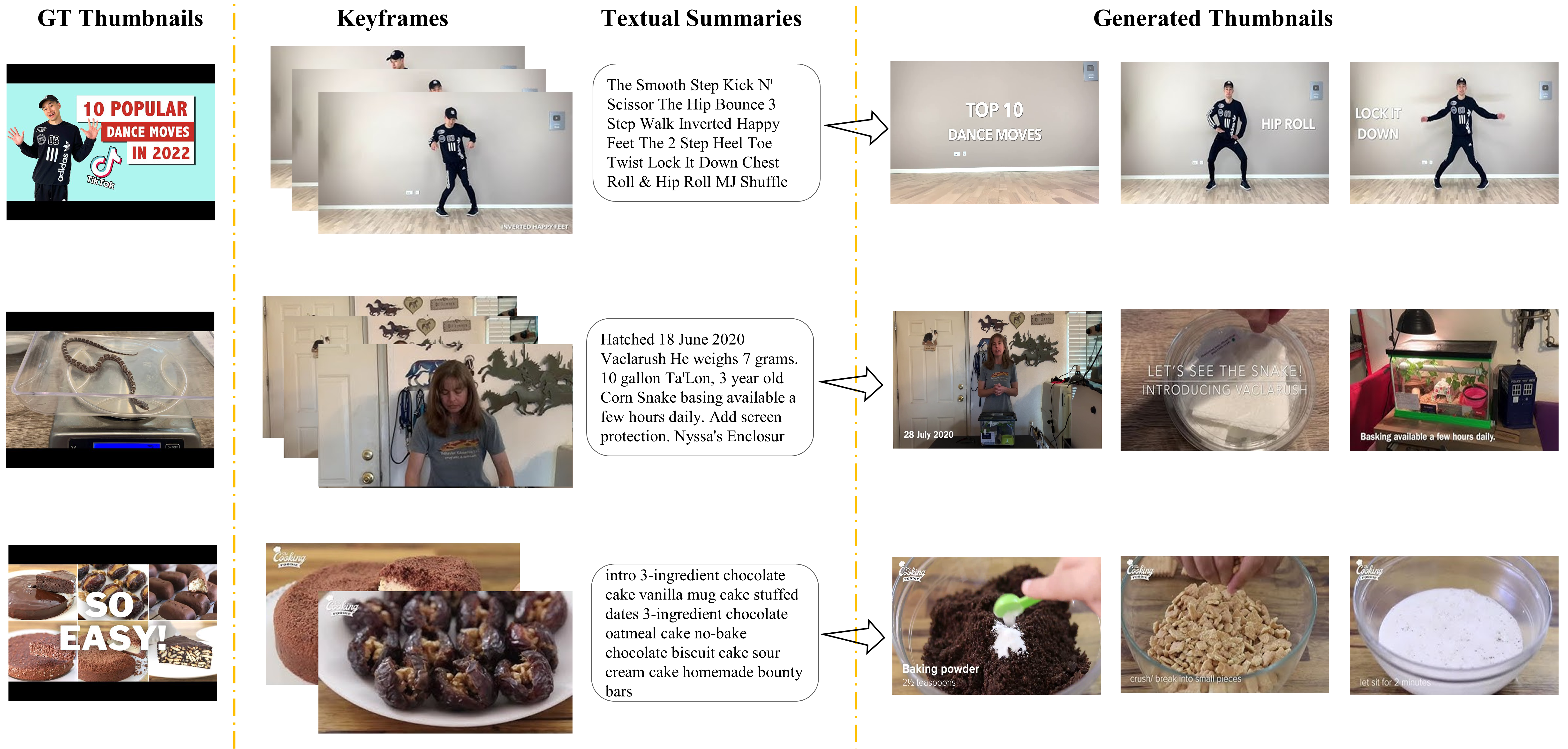}
		\caption{Comparison of GT thumbnails and our generated ones.}
		\label{fig:thumbnail}
		\vspace{-10pt}
	\end{figure*}

	\subsection{Results and Discussion}
	

	\paragraph{Supervised methods outperform unsupervised methods on video summarization}

	In our video summarization study, we have chosen the following methods as our baseline comparisons: Uniform Sampling \cite{jadon2019video}, K-means Clustering \cite{Hartigan1979AKC}, and VSUMM \cite{de2011vsumm}. The results, presented in Table~\ref{table:vs_results}, are under various evaluation metrics. For RMSE and SRE, lower values indicate better performance, whereas, for the remaining metrics, higher values are desirable. From Table~\ref{table:vs_results}, we can observe that VSUMM showcases the strongest performance among the baseline methods, yet it still falls short compared to our proposed method. But we can conclude that supervised methods outperform unsupervised methods.

	\paragraph{Pretrained large language models can still do well in text summarization}
	
	In the context of textual summarization, we have considered a set of representative models as our baseline comparisons: BERT2BERT \cite{turc2019}, BART \cite{Lewis2020BARTDS} (including BART-large-CNN and BART-large-XSUM), Distilbart \cite{Shleifer2020PretrainedSD}, T5 \cite{Raffel2019ExploringTL}, Pegasus \cite{Zhang2019PEGASUSPW}, and Longformer Encoder-Decoder (LED) \cite{Beltagy2020LongformerTL}. The performance of these models is summarized in Table~\ref{table:ts_results}. Among the baselines, T5, BART-large-XSUM, BART-large-CNN, and BERT2BERT exhibit superior performance, with T5 demonstrating relatively better results across various text evaluation metrics. In addition, the ROUGE score may not effectively capture performance differences compared to other evaluation metrics, because ROUGE does not take into account the semantic meaning and the factual accuracy of the summaries.

	\paragraph{MSMO results may depend on segmentation results and summarization methods}
	
	In the field of MSMO, we encountered limitations in accessing the codebases of existing works such as \cite{Zhu2018MSMOMS,hori2019end,zadeh-etal-2017-tensor,chen2018abstractive,Fu2021MMAVSAF,Fu2020MultimodalSF}. Therefore, we independently implemented several baselines to evaluate their performance on the MMSum dataset. For this purpose, we utilized LGSS as the segmentation backbone, VSUMM as the video summarizer, and selected text summarizers that exhibited the best performance in text summarization. The results are presented in Table~\ref{table:msmo_results}. Based on the findings, it is evident that the aforementioned combination approaches still fall short in comparison to our proposed method. This also indicates that the accuracy of temporal segmentation is crucial prior to generating summaries, highlighting it as a critical step and task preceding MSMO.


	\subsection{Thumbnail Generation}
	
	One direct and practical application of the MSMO task is to automatically generate thumbnails for a given video, which has become increasingly valuable in various real-world applications. With the exponential growth of online videos, effective and efficient methods are required to extract visually appealing and informative thumbnail representations. 
	In addition, many author-generated thumbnails involve words or titles that describe the whole video to attract more users. 
	In the context of online platforms, such as video-sharing websites or social media platforms, compelling thumbnails can significantly impact user engagement, content discoverability, and overall user experience. The benefits of automated thumbnail generation extend beyond user engagement and content discoverability. In e-commerce, for instance, thumbnails can play a vital role in attracting potential buyers by effectively showcasing products or services. Similarly, in video editing workflows, quick and accurate thumbnail generation can aid content creators in managing and organizing large video libraries efficiently.

	In our setting, we take advantage of the results by MSMO, which contain both visual summaries and text summaries, and combine them to generate thumbnails for a given video.
	In summary, the selected keyframes and generated textual summaries from the MSMO task are subsequently utilized to create the thumbnail. 
	To ensure an aesthetically pleasing appearance, we randomly sample from a corpus of fonts from Google Fonts and font sizes to utilize in the generated thumbnails. 
	Moreover, a random set of coordinates on the selected keyframe is sampled for the placement of the text. 
	Finally, the text is pasted onto the keyframe from the outputted set of coordinates to complete thumbnail generation. 
	
	More specifically, the font is randomly selected from 100 fonts, and the size of the font varies by 175 font sizes. Here we list 20 examples of fonts we used in our experiments: [Roboto, Open Sans, Lato, Montserrat, Raleway, Oswald, Source Sans Pro, Poppins, Noto Sans, Roboto Slab, Merriweather, Ubuntu, PT Sans, Playfair Display, Fira Sans, Nunito, Roboto Condensed, Zilla Slab, Arvo, Muli]. 
	We randomly select one font and a random font size. Given the image size of the selected keyframes, we also randomly select coordinates for where the text should be pasted onto the selected keyframes. We then paste the generated textual summary, which is modified by the randomly selected font and font size, onto the selected keyframes.
	Some examples are shown in Figure~\ref{fig:thumbnail}. More results can be found in Appendix~\ref{sec:appendix-thumbnail-results}.
	
	\vspace{-5pt}
	\paragraph{Limitations and Future Work Directions}
	The lack of publicly available MSMO baselines in existing literature underscores a significant gap, emphasizing the need for future efforts in this area. Advancing the field requires tackling the complex task of creating a diverse and extensive collection of baselines.
	
	Despite the progress made in automated thumbnail generation, challenges remain. These include enhancing the accuracy of thumbnail selection, accommodating various video genres and content types, and taking into account user preferences and context-specific requirements.
	
	Moreover, addressing ethical concerns related to potential biases, representation, and content moderation is crucial to ensuring fair and inclusive thumbnail generation. Exploring new quantitative evaluation metrics for the thumbnail generation task could also pave the way for valuable advancements in this domain.

	\section{Conclusion}

	In this research, our main goal was to overcome the limitations of existing MSMO datasets by creating a comprehensive dataset called MMSum. MMSum was meticulously curated to ensure top-notch quality of MSMO data, making it a valuable resource for tasks like video summarization, text summarization, and multimodal summarization.
	Additionally, we introduced a novel benchmark based on the MMSum dataset. This benchmark enables researchers and practitioners to assess their algorithms and models across a range of tasks.
	Moreover, leveraging the results from MSMO, we introduced a new task: automatically generating thumbnails for videos. This innovation has the potential to significantly enhance user engagement, content discoverability, and overall user experience. We hope that our MMSum dataset can contribute to the advancement of research in the MSMO field.

	\clearpage

	\clearpage
	{\small
		\bibliographystyle{ieee_fullname}
		\bibliography{egbib}
	}
	
	\clearpage
	\onecolumn
	\appendix
	\section{Datasheet}

These questions were copied from ``Datasheets for Datasets" \cite{Gebru2018DatasheetsFD}.

\subsection{Motivation}

The questions in this section are primarily intended to encourage
dataset creators to clearly articulate their reasons for creating the
dataset and to promote transparency about funding interests.
The latter may be particularly relevant for datasets created for
research purposes.\\

\begin{itemize}

\item \textbf{For what purpose was the dataset created?} Was there a specific task in mind? Was there a specific gap that needed to be filled? Please provide a description.

\textcolor{blue}{Multimodal summarization with multimodal output (MSMO) has emerged as a promising research direction. 
Nonetheless, numerous limitations exist within existing public MSMO datasets, including insufficient upkeep, data inaccessibility, limited size, and the absence of proper categorization, which pose significant challenges to effective research.
To address these challenges and provide a comprehensive dataset for this new direction, we have meticulously curated the \textbf{MultiSum} dataset. }

\item \textbf{Who created the dataset (e.g., which team, research group) and on behalf of which entity (e.g., company, institution, organization)?}

\textcolor{blue}{Our institution (will release the identity later).}

\item \textbf{Who funded the creation of the dataset?} If there is an associated grant, please provide the name of the grantor and the grant name and number.

\textcolor{blue}{Our institution (will release the identity later).}

\item \textbf{Any other comments?}

\textcolor{blue}{No. }

\end{itemize}

\subsection{Composition}

Dataset creators should read through \edit{these questions} prior to
any data collection and then provide answers once \edit{data} collection is
complete. Most of the questions \edit{in this section} are intended to
provide dataset consumers with the information they need to make
informed decisions about using the dataset for their chosen
tasks. Some of the questions are \edit{designed to elicit} information
about compliance with the EU's General Data Protection Regulation
(GDPR) or comparable regulations in other jurisdictions.

\edit{Questions that apply only to datasets that relate to people are
grouped together at the end of the section. We recommend taking a
broad interpretation of whether a dataset relates to people. For
example, any dataset containing text that was written by people
relates to people.}\\

\begin{itemize}

\item \textbf{What do the instances that comprise the dataset
    represent (e.g., documents, photos, people, countries)?} Are there
  multiple types of instances (e.g., movies, users, and ratings;
  people and interactions between them; nodes and edges)? Please
  provide a description.

  \textcolor{blue}{Videos, transcripts, keyframes, textual summaries, segmentation boundaries, titles, authors, and thumbnails. }

\item \textbf{How many instances are there in total (of each type, if appropriate)?}

\textcolor{blue}{17 main categories and 170 subcategories, 5,100 videos, in a total of 1229.9 hours. }

\item \textbf{Does the dataset contain all possible instances or is it
    a sample (not necessarily random) of instances from a larger set?}
  If the dataset is a sample, then what is the larger set? Is the
  sample representative of the larger set (e.g., geographic coverage)?
  If so, please describe how this representativeness was
  validated/verified. If it is not representative of the larger set,
  please describe why not (e.g., to cover a more diverse range of
  instances, because instances were withheld or unavailable).

   \textcolor{blue}{It contains all possible instances.}

\item \textbf{What data does each instance consist of?} ``Raw'' data
  (e.g., unprocessed text or images) or features? In either case,
  please provide a description.

  \textcolor{blue}{Video, transcripts, keyframes, textual summaries, segmentation boundaries, titles, authors, and thumbnails.}

\item \textbf{Is there a label or target associated with each
    instance?} If so, please provide a description.

    \textcolor{blue}{Segmentation boundaries are labels for video temporal segmentation.  Keyframes and textual summaries are labels for MSMO. Thumbnails are the ground-truth for thumbnail generation. }

\item \textbf{Is any information missing from individual instances?}
  If so, please provide a description, explaining why this information
  is missing (e.g., because it was unavailable). This does not include
  intentionally removed information, but might include, e.g., redacted
  text.

 \textcolor{blue}{ No. }

\item \textbf{Are relationships between individual instances made
    explicit (e.g., users' movie ratings, social network links)?} If
  so, please describe how these relationships are made explicit.

 \textcolor{blue}{N/A. }

\item \textbf{Are there recommended data splits (e.g., training,
    development/validation, testing)?} If so, please provide a
  description of these splits, explaining the rationale behind them.

  \textcolor{blue}{Yes, for the train/val/test split, since our dataset is already randomly collected from YouTube, we designate the last 30\% of videos within each subcategory (indexed 21-29) as the testing set. The remaining videos are then assigned to the training set (indexed 00-20) in each subcategory.}

\item \textbf{Are there any errors, sources of noise, or redundancies
    in the dataset?} If so, please provide a description.

    \textcolor{blue}{No, but we would expect subjectivity exists in the annotations. }

\item \textbf{Is the dataset self-contained, or does it link to or
    otherwise rely on external resources (e.g., websites, tweets,
    other datasets)?} If it links to or relies on external resources,
    a) are there guarantees that they will exist, and remain constant,
    over time; b) are there official archival versions of the complete
    dataset (i.e., including the external resources as they existed at
    the time the dataset was created); c) are there any restrictions
    (e.g., licenses, fees) associated with any of the external
    resources that might apply to a \edit{dataset consumer}? Please provide
    descriptions of all external resources and any restrictions
    associated with them, as well as links or other access points, as
    appropriate.

    \textcolor{blue}{The Youtube links to the videos are provided in the public version. If some of the data become unavailable, we can provide the features for those videos. }

\item \textbf{Does the dataset contain data that might be considered
    confidential (e.g., data that is protected by legal privilege or
    by doctor\edit{--}patient confidentiality, data that includes the content
    of individuals' non-public communications)?} If so, please provide
    a description.

    \textcolor{blue}{No. }

\item \textbf{Does the dataset contain data that, if viewed directly,
    might be offensive, insulting, threatening, or might otherwise
    cause anxiety?} If so, please describe why.

    \textcolor{blue}{No. }

\end{itemize}

\edit{If the dataset does not }relate to people, you may skip the remaining questions in this section.

\begin{itemize}

\item \textbf{Does the dataset identify any subpopulations (e.g., by
    age, gender)?} If so, please describe how these subpopulations are
  identified and provide a description of their respective
  distributions within the dataset.

  \textcolor{blue}{No. }

\item \textbf{Is it possible to identify individuals (i.e., one or
    more natural persons), either directly or indirectly (i.e., in
    combination with other data) from the dataset?} If so, please
    describe how.

   \textcolor{blue}{No, too many people involved in the videos.}

\item \textbf{Does the dataset contain data that might be considered
    sensitive in any way (e.g., data that reveals rac\edit{e} or ethnic
    origins, sexual orientations, religious beliefs, political
    opinions or union memberships, or locations; financial or health
    data; biometric or genetic data; forms of government
    identification, such as social security numbers; criminal
    history)?} If so, please provide a description.

    \textcolor{blue}{No. }

\item \textbf{Any other comments?}

\textcolor{blue}{No. }

\end{itemize}

\subsection{Collection Process}

As with the \edit{questions in the} previous section, dataset creators should
read through these questions prior to any data collection to flag
potential issues and then provide answers once collection is complete.
\edit{In addition to the goals outlined in the previous section, the
questions in this section are designed to elicit information that may
help researchers and practitioners to create alternative datasets with
similar characteristics. Again, questions that apply only to datasets
that relate to people are grouped together at the end of the
section.}\\

\begin{itemize}

\item \textbf{How was the data associated with each instance
    acquired?} Was the data directly observable (e.g., raw text, movie
  ratings), reported by subjects (e.g., survey responses), or
  indirectly inferred/derived from other data (e.g., part-of-speech
  tags, model-based guesses for age or language)? If \edit{the} data was reported
  by subjects or indirectly inferred/derived from other data, was the
  data validated/verified? If so, please describe how.

  \textcolor{blue}{Data resources are publicly available online. }

\item \textbf{What mechanisms or procedures were used to collect the
    data (e.g., hardware apparatus\edit{es} or sensor\edit{s}, manual human
    curation, software program\edit{s}, software API\edit{s})?} How were these
    mechanisms or procedures validated?

  \textcolor{blue}{We open-sourced our data collection tool.}

\item \textbf{If the dataset is a sample from a larger set, what was
    the sampling strategy (e.g., deterministic, probabilistic with
    specific sampling probabilities)?}

    \textcolor{blue}{N/A. }

\item \textbf{Who was involved in the data collection process (e.g.,
    students, crowdworkers, contractors) and how were they compensated
    (e.g., how much were crowdworkers paid)?}

  \textcolor{blue}{Students and crowdworkers, and were rewarded with virtual currency. }

\item \textbf{Over what timeframe was the data collected?} Does this
  timeframe match the creation timeframe of the data associated with
  the instances (e.g., recent crawl of old news articles)?  If not,
  please describe the timeframe in which the data associated with the
  instances was created.

  \textcolor{blue}{No specific timeframe is set.}

\item \textbf{Were any ethical review processes conducted (e.g., by an
    institutional review board)?} If so, please provide a description
  of these review processes, including the outcomes, as well as a link
  or other access point to any supporting documentation.

  \textcolor{blue}{Yes. The legal counsel reviewed it.}

\end{itemize}

\edit{If the dataset does not relate to people, you may skip the remaining questions in this section.}

\begin{itemize}

\item \textbf{Did you collect the data from the individuals in
    question directly, or obtain it via third parties or other sources
    (e.g., websites)?}

    \textcolor{blue}{From YouTube.}

\item \textbf{Were the individuals in question notified about the data
    collection?} If so, please describe (or show with screenshots or
  other information) how notice was provided, and provide a link or
  other access point to, or otherwise reproduce, the exact language of
  the notification itself.

  \textcolor{blue}{N/A.}

\item \textbf{Did the individuals in question consent to the
    collection and use of their data?} If so, please describe (or show
  with screenshots or other information) how consent was requested and
  provided, and provide a link or other access point to, or otherwise
  reproduce, the exact language to which the individuals consented.

  \textcolor{blue}{We follow the license from YouTube.}

\item \textbf{If consent was obtained, were the consenting individuals
    provided with a mechanism to revoke their consent in the future or
    for certain uses?} If so, please provide a description, as well as
  a link or other access point to the mechanism (if appropriate).

  \textcolor{blue}{We follow the license from YouTube.}

\item \textbf{Has an analysis of the potential impact of the dataset
    and its use on data subjects (e.g., a data protection impact
    analysis) been conducted?} If so, please provide a description of
  this analysis, including the outcomes, as well as a link or other
  access point to any supporting documentation.

  \textcolor{blue}{No. }

\item \textbf{Any other comments?}

\textcolor{blue}{No. }

\end{itemize}

\subsection{Preprocessing/cleaning/labeling}

Dataset creators should read through these questions prior to any
preprocessing, cleaning, or labeling and then provide answers once
these tasks are complete. The questions in this section are intended
to provide dataset consumers with the information they need to
determine whether the ``raw'' data has been processed in ways that are
compatible with their chosen tasks. For example, text that has been
converted into a ``bag-of-words'' is not suitable for tasks involving
word order.\\

\begin{itemize}

\item \textbf{Was any preprocessing/cleaning/labeling of the data done
    (e.g., discretization or bucketing, tokenization, part-of-speech
    tagging, SIFT feature extraction, removal of instances, processing
    of missing values)?} If so, please provide a description. If not,
  you may skip the remain\edit{ing} questions in this section.

  \textcolor{blue}{No. }

\item \textbf{Was the ``raw'' data saved in addition to the preprocessed/cleaned/labeled data (e.g., to support unanticipated future uses)?} If so, please provide a link or other access point to the ``raw'' data.

\textcolor{blue}{The dataset will be available on our website.}

\item \textbf{Is the software \edit{that was} used to preprocess/clean/label the \edit{data} available?} If so, please provide a link or other access point.

\textcolor{blue}{Yes, the data collection tool will be available on our website.}

\item \textbf{Any other comments?}

\textcolor{blue}{No. }

\end{itemize}

\subsection{Uses}

\edit{The} questions \edit{in this section} are intended to encourage dataset
creators to reflect on the tasks for which the dataset should and
should not be used. By explicitly highlighting these tasks, dataset
creators can help dataset consumers to make informed decisions,
thereby avoiding potential risks or harms.\\

\begin{itemize}

\item \textbf{Has the dataset been used for any tasks already?} If so, please provide a description.

\textcolor{blue}{Yes, the dataset has been used for a series of tasks, including video temporal segmentation, video summarization, text summarization, multimodal summarization (MSMO), and thumbnail generation.}

\item \textbf{Is there a repository that links to any or all papers or systems that use the dataset?} If so, please provide a link or other access point.

\textcolor{blue}{Yes, it will be released on our website.}

\item \textbf{What (other) tasks could the dataset be used for?}

\textcolor{blue}{Video related, multimodal related.}

\item \textbf{Is there anything about the composition of the dataset or the way it was collected and preprocessed/cleaned/labeled that might impact future uses?} For example, is there anything that a \edit{dataset consumer} might need to know to avoid uses that could result in unfair treatment of individuals or groups (e.g., stereotyping, quality of service issues) or other \edit{risks or} harms (e.g., \edit{legal risks,} financial harms\edit{)?} If so, please provide a description. Is there anything a \edit{dataset consumer} could do to mitigate these \edit{risks or} harms?

\textcolor{blue}{No. }

\item \textbf{Are there tasks for which the dataset should not be used?} If so, please provide a description.

\textcolor{blue}{No. }

\item \textbf{Any other comments?}

\textcolor{blue}{No. }

\end{itemize}

\subsection{Distribution}

Dataset creators should provide answers to these questions prior to
distributing the dataset either internally within the entity on behalf
of which the dataset was created or externally to third parties.\\

\begin{itemize}

\item \textbf{Will the dataset be distributed to third parties outside of the entity (e.g., company, institution, organization) on behalf of which the dataset was created?} If so, please provide a description.

\textcolor{blue}{No. }

\item \textbf{How will the dataset will be distributed (e.g., tarball on website, API, GitHub)?} Does the dataset have a digital object identifier (DOI)?

\textcolor{blue}{Self-contained website.}

\item \textbf{When will the dataset be distributed?}

\textcolor{blue}{Will be released soon.}

\item \textbf{Will the dataset be distributed under a copyright or other intellectual property (IP) license, and/or under applicable terms of use (ToU)?} If so, please describe this license and/or ToU, and provide a link or other access point to, or otherwise reproduce, any relevant licensing terms or ToU, as well as any fees associated with these restrictions.

\textcolor{blue}{CC BY-NC-SA License.}

\item \textbf{Have any third parties imposed IP-based or other restrictions on the data associated with the instances?} If so, please describe these restrictions, and provide a link or other access point to, or otherwise reproduce, any relevant licensing terms, as well as any fees associated with these restrictions.

\textcolor{blue}{No.}

\item \textbf{Do any export controls or other regulatory restrictions apply to the dataset or to individual instances?} If so, please describe these restrictions, and provide a link or other access point to, or otherwise reproduce, any supporting documentation.

\textcolor{blue}{No.}

\item \textbf{Any other comments?}

\textcolor{blue}{No.}

\end{itemize}

\subsection{Maintenance}

As with the \edit{questions in the} previous section, dataset creators
should provide answers to these questions prior to distributing the
dataset. The questions \edit{in this section} are intended to
encourage dataset creators to plan for dataset maintenance and
communicate this plan \edit{to} dataset consumers.\\

\begin{itemize}

\item \textbf{Who \edit{will be} supporting/hosting/maintaining the dataset?}

\textcolor{blue}{Our institute.}

\item \textbf{How can the owner/curator/manager of the dataset be contacted (e.g., email address)?}

\textcolor{blue}{We will provide the contact information on the webpage.}

\item \textbf{Is there an erratum?} If so, please provide a link or other access point.

\textcolor{blue}{We will provide the contact information on the webpage.}

\item \textbf{Will the dataset be updated (e.g., to correct labeling
    errors, add new instances, delete instances)?} If so, please
  describe how often, by whom, and how updates will be communicated to
  \edit{dataset consumers} (e.g., mailing list, GitHub)?

  \textcolor{blue}{Currently no, but we will post an announcement on the website once there is a new version available. }

\item \textbf{If the dataset relates to people, are there applicable
    limits on the retention of the data associated with the instances
    (e.g., were \edit{the} individuals in question told that their data would
    \edit{be} retained for a fixed period of time and then deleted)?} If so,
    please describe these limits and explain how they will be
    enforced.

    \textcolor{blue}{No. }

\item \textbf{Will older versions of the dataset continue to be
    supported/hosted/maintained?} If so, please describe how. If not,
  please describe how its obsolescence will be communicated to \edit{dataset
  consumers}.

  \textcolor{blue}{Yes. All the released versions will be hosted on the same website.}

\item \textbf{If others want to extend/augment/build on/contribute to
    the dataset, is there a mechanism for them to do so?} If so,
  please provide a description. Will these contributions be
  validated/verified? If so, please describe how. If not, why not? Is
  there a process for communicating/distributing these contributions
  to \edit{dataset consumers}? If so, please provide a description.

  \textcolor{blue}{Under the same license and contact our team.}

\item \textbf{Any other comments?}

\textcolor{blue}{No. }

\end{itemize}

\subsection{URL to Website}

\textcolor{blue}{Will be available soon.}

\subsection{Responsibility Statement}

\textcolor{blue}{The authors declare that they bear all responsibility for violations of rights related to this dataset.}

\subsection{Hosting, licensing, and maintenance plan}

\textcolor{blue}{The dataset will be properly hosted and maintained through our website. The dataset is under CC BY-NC-SA License. }

\subsection{Data Format}

\textcolor{blue}{The released dataset includes (1) the annotation files in the $.json$ format, including video transcripts, video segmentation boundaries, textual summaries for each segment, titles, authors, and timestamps.  (2) the keyframe files in the $.png$ format. (3) The thumbnail files in the $.png$ format.}

\subsection{Long-term preservation}

\textcolor{blue}{The authors guarantee that the dataset will be properly hosted and maintained through our website for a long time. }

\section{Categories of MMSum Dataset}\label{sec:appendix_categories}

\begin{figure*}[htp]
  \centering
  \includegraphics[width=0.99\linewidth]{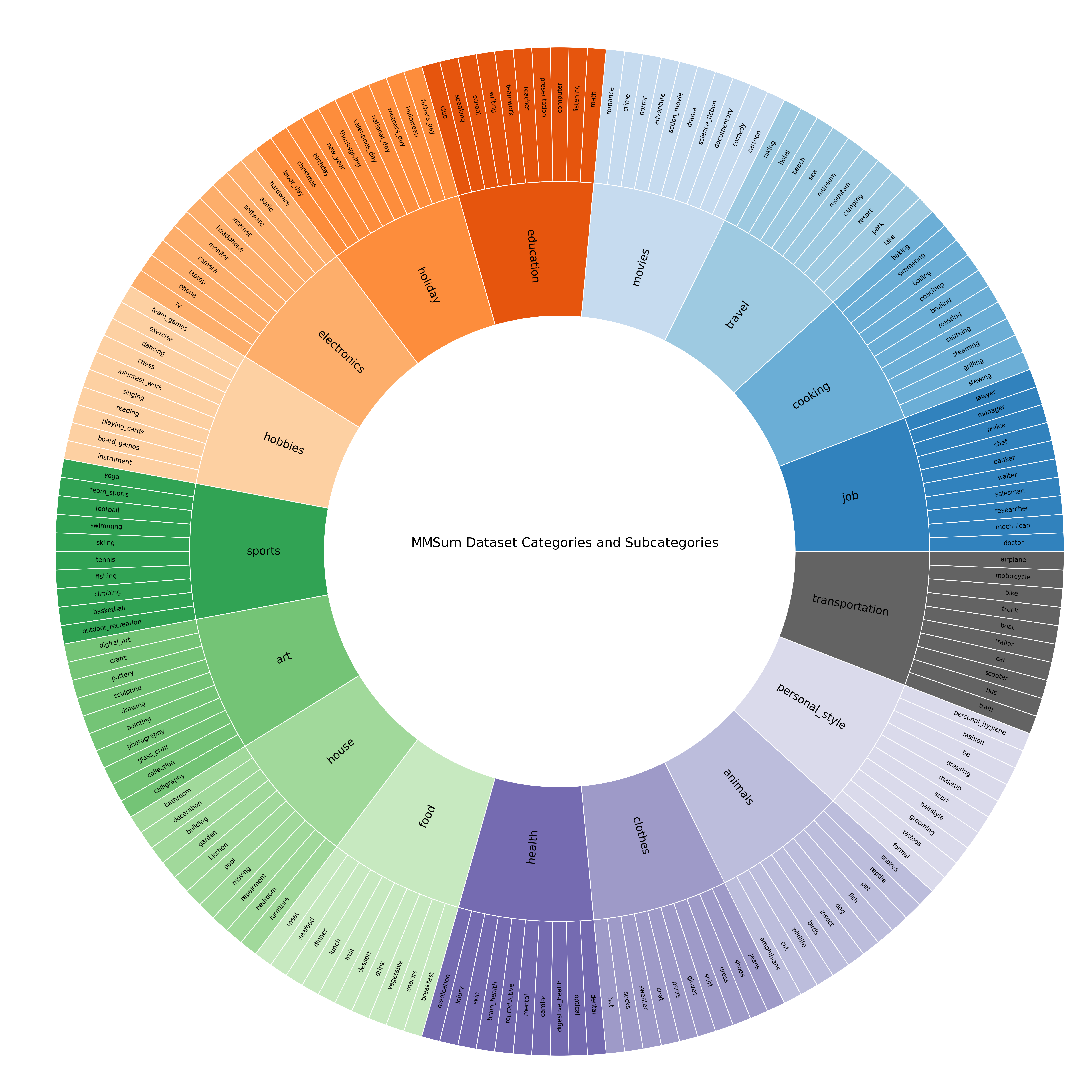}
  \caption{The 17 categories (with 10 subcategories each) of the MMSum dataset, resulting in 170 subcategories in total.}
  \label{fig:appendix_categories}
\end{figure*}

\begin{figure*}[htp]
  \centering
  \includegraphics[width=0.7\linewidth]{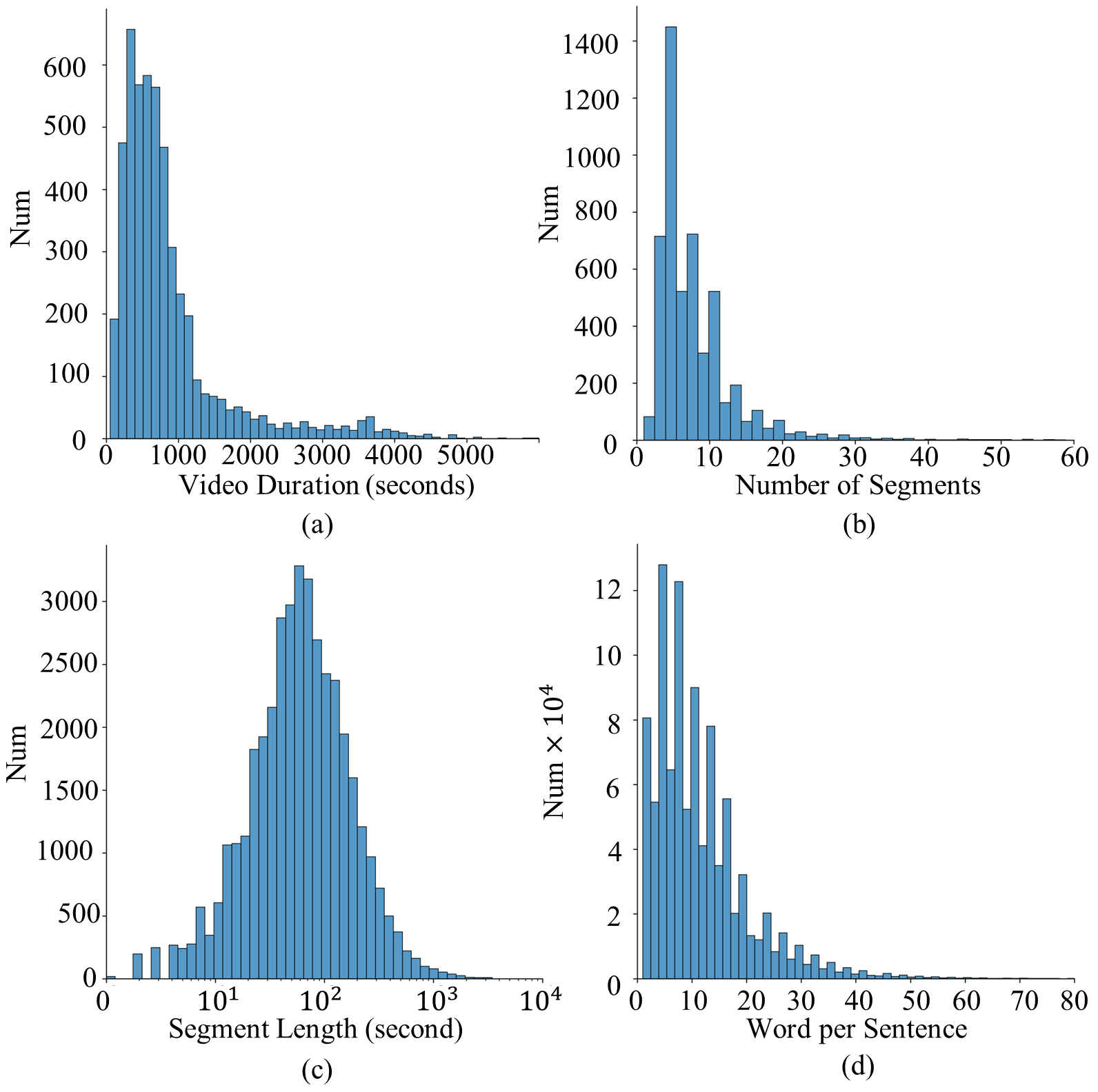}
  \caption{The statistics of the MMSum dataset, which show the distribution of (a) video duration; (b) number of segments per video; (c) segment duration; (d) number of words per sentence}
\end{figure*}

\section{Tasks}\label{sec:tasks}

Our dataset contains sufficient information, making it possible to conduct many downstream tasks, such as video temporal segmentation (VTS), video summarization (VS), text summarization (TS), and multimodal video summarization with multimodal output (MSMO). To make it more clear, we highlight the description of each task and the differences between them.

\vspace{-2pt}
\paragraph{Video Temporal Segmentation (VTS)} 
Video temporal segmentation (VTS) is the process of partitioning a video sequence into disjoint sets of consecutive frames that are homogeneous according to some defined criteria. Normally, VTS aims at splitting the whole video into several small segments based on video scene change, which is also related to video shot detection and video transition detection. Multimodal Video Temporal Segmentation (M-VTS) differs from VTS, where textual data (video transcript) is also used as inputs for splitting the input video into small video segments.

\vspace{-2pt}
\paragraph{Video Summarization (VS)} 
Video Summarization aims at generating a short synopsis that summarizes the video content by selecting the most informative and vital parts. The input only contains visual information and uses computer vision mechanisms to generate summaries.

\vspace{-2pt}
\paragraph{Text Summarization (TS)} 
Textual summarization takes textual metadata, i.e., documents, articles, tweets, etc, as input, and generates textual summaries, in two directions: abstractive summarization and extractive summarization. Abstractive methods select words based on semantic understanding, and even the words may not appear in the source \cite{tan2017abstractive,Abigail2017}. Extractive methods attempt to summarize language by selecting a subset of words that retain the most critical points, which weights the essential part of sentences to form the summary \cite{narayan2018ranking, wu2018learning}. 
\paragraph{Multimodal Summarization with Multimodal Output (MSMO)}
MSMO aims to produce both visual and textual summaries for a given video. Different from pure video summarization, MSMO takes both visual and textual information as inputs and outputs both visual and textual summaries.

\begin{table}[tp]
\centering
\caption{Categories and sub-categories for the MMSum dataset.}
\begin{adjustbox}{width=0.9\linewidth}
 \begin{tabular}{lp{13cm}r}
    \toprule
      Category & Sub-categories  & Number   \\ 
     \midrule
     Animals & Dog, Wildlife, Cat, Fish, Birds, Insect, Snakes, Pet, Amphibians, Reptile  & 30 $\times$ 10 =300 \\
     Education & School, Club, Teacher, Speaking, Listening, Writing, Presentation, Math, Computer, Teamwork & 30 $\times$ 10 =300 \\
     Health & Mental, Injury, Medication, Digestive health, Dental, Optical, Reproductive, Skin, Brain health, Cardiac & 30 $\times$ 10 =300 \\
     Travel &Museum, Park, Sea, Beach, Mountain, Lake, Hotel, Resort, Camping, Hiking & 30 $\times$ 10 =300 \\
     Movies &Action movie, Comedy, Romance, Science fiction, Horror, Drama, Cartoon,  Documentary, Adventure, Crime & 30 $\times$ 10 =300 \\
     Cooking & Broiling, Grilling, Roasting, Baking, Sauteing, Boiling, Steaming, Poaching, Simmering, Stewing & 30 $\times$ 10 =300 \\
     Job &Manager, Researcher, Chef, Police, Lawyer, Salesman, Mechnican, Banker, Doctor, Waiter & 30 $\times$ 10 =300 \\
     Electronics & laptop, TV, Phone, Software, Internet, Camera, Audio, Headphone, Hardware, Monitor & 30 $\times$ 10 =300 \\ 
     Art & Crafts, Photography, Painting, Collection, Drawing, Digital art, sculpting, pottery, glass craft, calligraphy & 30 $\times$ 10 =300 \\ 
     Personal Style &Grooming, Fashion, Personal Hygiene, Tattoos, Scarf, Hair Style, Makeup, Dressing, Tie, Formal    & 30 $\times$ 10 =300 \\
     Clothes & Sweater, Jeans, Shirt, Socks, Coat, Pants, Hat, Gloves, Dress, Shoes & 30 $\times$ 10 =300 \\ 
     Sports & Outdoor recreation, Team sports, Tennis, Football, Basketball, Climbing, Skiing, Swimming, Fishing, Yoga & 30 $\times$ 10 =300 \\ 
     House & Building, Garden, Pool, Bathroom, Bedroom, Kitchen, Repairment, Moving, Decoration, Furniture & 30 $\times$ 10 =300 \\ 
     Food & Fruit, Vegetable, Drink, Meat, Seafood, Snacks, Dessert, Breakfast, Lunch, Dinner & 30 $\times$ 10 =300\\
     Holiday & Halloween, Christmas, Labor day, Thanksgiving, Valentines day, Mother's day, Birthday, National day, New year, Father's day & 30 $\times$ 10 =300\\
     Transportation & Car, Train, Bus, Boat, Bike, Airplane, Motorcycle, Truck, Trailer, Scooter & 30 $\times$ 10 =300 \\
     Hobbies & Dancing, Singing, Playing cards, Reading, Chess, Board games, Team games, Volunteer work, Instrument, Exercise & 30 $\times$ 10 =300 \\
     \midrule
     Total &---------------------------------------------------------------------------------------- &17$\times$30$\times$10=5,100 \\
    \bottomrule
\end{tabular}  
\end{adjustbox}
\label{Table:category}
\vspace{-5pt}
\end{table}

\section{More Details about Our Model}\label{sec:appendix-model}
\vspace{-2pt}
\paragraph{Text Encoder} The Transformer encoder \cite{Vaswani2017AttentionIA} is employed to convert the text into a sequence of token embeddings. Inspired by \cite{Yu2021VisionGG,Krubiski2023MLASKMS}, we initialize the encoder's weights using the pre-trained mT5 model \cite{Xue2020mT5AM}. To investigate the impact of task-specific pre-training, we fine-tune mT5 on the text-to-text summarization task, where $X_{e n c}=\operatorname{TextEncoder}(X)$. 

\vspace{-2pt}
\paragraph{Video Encoder} To capture short-term temporal dependencies, we utilize 3D convolutional networks as in \cite{Krubiski2023MLASKMS}. We partition the video into non-overlapping frame sequences and employ a 3D CNN network for feature extraction. Specifically, we utilize two different feature extractors. Firstly, we utilize the $\mathrm{R}(2+1) \mathrm{D}$ model trained by \cite{Ghadiyaram2019LargeScaleWP} for video action recognition on weakly-supervised social-media videos. Secondly, we utilize the visual component of the S3D Text-Video model trained in a self-supervised manner by \cite{Miech2019EndtoEndLO} on the HowTo100M dataset \cite{Miech2019HowTo100MLA}.
To incorporate long-term temporal dependencies, we process the sequence of video features using a Transformer encoder. This enables us to effectively capture and model the relationships between video frames over an extended duration, where 
$V_{\text {enc }}=3 \mathrm{D}-\mathrm{CNN}(V), 
V_{e n c}=\operatorname{VideorEncoder}\left(V_{\text {enc }}\right)$.

\begin{figure}[t]
  \centering
  \includegraphics[width=0.99\linewidth]{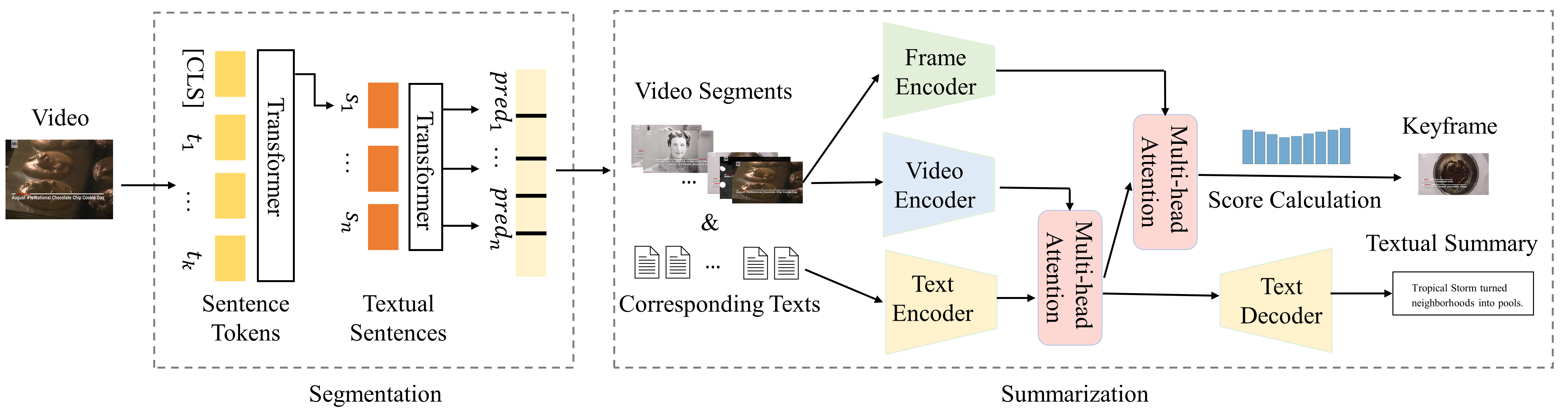}
  \vspace{-5pt}
  \caption{An overview of our model.}
  \label{fig:model_appendix}
  \vspace{-10pt}
\end{figure}

\vspace{-2pt}
\paragraph{Frame Encoder} To facilitate the selection of a specific frame as a cover picture, we require frame-level representations \cite{Krubiski2023MLASKMS}. In our experimental setup, we sample one frame per second from the video. For feature extraction, we employ two models: EfficientNet \cite{Tan2019EfficientNetRM} and Vision Transformer (ViT) \cite{Dosovitskiy2020AnII}. Both models were pre-trained on the ImageNet dataset \cite{Russakovsky2014ImageNetLS} for image classification tasks.
To provide contextual information, we process the sequence of frame features using a Transformer encoder, which captures the relationships and dependencies between the frame-level representations, enabling a more comprehensive understanding of the video content. Before applying the Transformer encoder, we ensure that both the video features and frame features have the same dimensions as the hidden states of the text encoder. In the case of a single model, the two sets of features are concatenated together before undergoing the projection step.
\begin{equation}
\begin{gathered}
V_{\text {frame }}=\mathrm{CNN}(\operatorname{Sample}(V)), 
V_{\text {frame }}=\operatorname{FrameEncoder}\left(V_{\text {frame }}\right)
\end{gathered}
\end{equation}

\paragraph{Multi-head Attention}

In line with the study conducted by \cite{Yu2021VisionGG,Krubiski2023MLASKMS}, which explored various methods of integrating visual information into pre-trained generative language models, we adopt the approach of multi-head attention-based fusion. This technique allows us to obtain a vision-guided text representation by incorporating visual information into the model. The fusion process takes place after the last encoder layer, ensuring that both textual and visual inputs are combined effectively to enhance the overall representation.

\begin{equation}
\begin{aligned}
& Q=X_{e n c} W_q, Q \in \mathbb{R}^{M \times d},
 K=V_{e n c} W_k, K \in \mathbb{R}^{N^{\prime} \times d} \\
& V=V_{e n c} W_v, V \in \mathbb{R}^{N^{\prime} \times d},
 \widetilde{X}_{e n c}= \operatorname{MHA}(Q, K, V), \widetilde{X}_{e n c} \in \mathbb{R}^{M \times d}
\end{aligned}
\end{equation}

As recommended by \cite{Liu2020MultistageFW,Krubiski2023MLASKMS}, we incorporate the use of the forget gate mechanism (FG) in our model. This mechanism enables the model to filter out low-level cross-modal adaptation information. By utilizing the forget gate, our model can selectively retain and focus on the most relevant and informative features, disregarding less important or noisy information during the cross-modal fusion process. This helps improve the overall performance and robustness of the model in handling multimodal data.
\begin{equation}\small
\widehat{X}_{\text {enc }}=\operatorname{FG}\left(X_{\text {enc }}, \widetilde{X}_{\text {enc }}\right), \widehat{X}_{\text {enc }} \in \mathbb{R}^{M \times d}
\end{equation}

To obtain the text+video guided frame representations, we employ the same multi-head attention mechanism. However, in this case, we substitute the input $X_{enc}$ with $V_{frame}$ and $V_{enc}$ with $\widehat{X}{enc}$. By using the video frame features $V{frame}$ and the transformed text representations $\widehat{X}{enc}$, we generate the guided frame representations $\widehat{V}{frame}$ through the multi-head attention process. This allows us to effectively incorporate both textual and visual information, guiding the frame-level representations based on the context provided by the text and video.

\paragraph{Text Decoder}

To generate the textual summary, we employ a standard Transformer decoder, initializing its weights with the mT5 checkpoint. The vision-guided text representation $\widehat{X}_{enc}$ serves as the input to the decoder. During training, we utilize the standard negative log-likelihood loss (NLLLoss) with respect to the target sequence $Y$. This loss function measures the dissimilarity between the predicted summary generated by the model and the ground truth summary, allowing the model to learn and improve its summary generation capabilities through backpropagation.

\begin{equation}\small
\begin{gathered}
\widehat{Y}=\text { TransformerDecoder }\left(\widehat{X}_{e n c}\right),
\mathcal{L}_{\text {text }}=\text { NLLLoss }(\widehat{Y}, Y)
\end{gathered}
\end{equation}

To obtain the labels $C$ for the cover picture (cover frame) selection, we calculate the cosine similarity between the CNN features of the reference cover picture and the candidate frames. In most instances, the similarity values fall within the range of [0, 1], while the remaining negative values are mapped to 0. Previous studies such as \cite{li2020vmsmo} and \cite{Fu2020MultimodalSF} considered the frame with the maximum cosine similarity as the ground truth (denoted as $C_{\max}$), while considering the other frames as negative samples.
However, upon analyzing the cosine similarity patterns, we observed that some videos exhibit multiple peaks or consecutive sequences of frames with very similar scores, capturing still scenes. We recognized that this could potentially harm the model's performance, as very similar frames might be labeled as both positive and negative examples. To address this issue, in addition to the binary labels $C_{\max}$, we introduce smooth labels denoted as $C_{\text{smooth}}$. These smooth labels assign to each frame its cosine similarity score with the reference cover picture. By incorporating the smooth labels, we aim to provide a more nuanced and continuous representation of the frame similarities, allowing the model to learn from a broader range of similarity scores during the training process.

In our approach, we utilize a projection matrix to map the text+video guided frame representations $\widehat{V}_{frame}$ to a single dimension. This dimension reduction step allows us to obtain a compact representation of the frame features. Subsequently, we train the model using the binary cross-entropy (CE) loss, where the target labels $C$ can either be $C{\max}$ or $C_{\text{smooth}}$.
To train the entire model in an end-to-end fashion, we minimize the sum of losses $\mathcal{L}$, which includes the negative log-likelihood loss for textual summary generation and the binary cross-entropy loss for cover picture selection. By jointly optimizing these losses, the model learns to generate accurate summaries and make effective cover picture selections based on the input text and video.
Please note that $\mathcal{L}$ refers to the combined loss function that encompasses both the negative log-likelihood loss for summary generation and the binary cross-entropy loss for cover picture selection.
\vspace{-2pt}
\begin{equation}\small
\begin{gathered}
\widehat{C}=\widehat{V}_{\text {frame }} W_p, W_p \in \mathbb{R}^{d \times 1} ,
\mathcal{L}_{\text {image }}=\mathrm{CE}(\widehat{C}, C),
\mathcal{L}=\mathcal{L}_{\text {text }}+\mathcal{L}_{\text {image }}
\end{gathered}
\end{equation}

\section{Baseline Implementation Details}\label{sec:appendix-baselines}

\vspace{-2pt}
\subsection{Video Temporal Segmentation} 

\paragraph{Video Temporal Segmentation Evaluation} For VTS, we followed \cite{Rao2020ALA} and adopted four common metrics: (1) Average Precision (AP); (2) F1 score; (3) $M_{iou}$: a weighted sum of the intersection of the union of a detected scene boundary with respect to its distance to the closest ground-truth scene boundary; and (4) Recall@$k$s: recall at $k$ seconds ($k = \{3,5,10\}$), the percentage of annotated scene boundaries which lies within $k$-second window of the predicted boundary.

The performance of video temporal segmentation has a great impact on the final performance, so in this section, we compare the performance of VTS with several baselines:  Histogram Intersect \cite{Lee2005EvaluationOI}, Moment Invariant \cite{Huang2010AnalysisOH}, Twin Comparison \cite{Zhang1993AutomaticPO}, PySceneDetect \cite{PySceneDetect}, and LGSS \cite{Rao2020ALA}.

\vspace{-2pt}
\paragraph{Histogram Intersect} 
We predict video boundaries at time $t$ when the overlap of color histograms in consecutive frames of the video 
\begin{equation}\small
    \frac{\sum_{b} \min(H_{t,b},H_{t-1,b})}{\sum H_{t-1}} \ge 0.5
\end{equation}
As in the original work \cite{Lee2005EvaluationOI}, we weighted the H, S, and V channels of the base image $0.5, 0.3, 0.2$ when constructing the histogram.

\vspace{-2pt}
\paragraph{Moment Invariant} 
We predict video boundaries at time $t$ when the distance between the Hu image moments of consecutive frames of the video $dist_{Hu}(I_{t}, I_{t-1}) \ge 0.3$.

\vspace{-2pt}
\paragraph{Twin Comparison} 
We define hyperparameters $T_s = 16, T_b = 3750$, such that the algorithm predicts the start of a segment at time $t$ where the difference between consecutive frames $D_{t,t-1} > T_s$, and the end of a segment at $t'$ when $D_{t,t'} > T_b$.

\vspace{-2pt}
\paragraph{PySceneDetect} 
We run the tool with hyperparameters $adaptive\_threshold = 64, min\_scene\_length = 5$.

\vspace{-2pt}
\paragraph{LGSS} 
We identify boundaries where the mean difference across channels H, S, and V between consecutive frames of the video $D_{HSV} \ge 20$ \cite{Rao2020ALA}.

\vspace{-2pt}
\subsection{Video Summarization}

For video summarization, we selected the following representative methods as our baselines: Uniform Sampling \cite{jadon2019video}, K-means Clustering \cite{Hartigan1979AKC}, Scale Invariant Feature Transform (SIFT) \cite{Morel2010IsT}, VSUMM \cite{de2011vsumm}, and Keyframe Extraction \cite{jadon2019video}.

\vspace{-2pt}
\paragraph{Uniform Sampling} We downsample the videos to 1 frame per second before taking 5 percent of the video frames, evenly spacing them throughout the video to have a uniform sample of key frames.

\vspace{-2pt}
\paragraph{K-means Clustering}  We compute the video's histogram per frame and apply K-means to find relevant frames for the summarization process. To extract the required images, images were captured at 1 FPS using the cv2 library.  

\vspace{-2pt}
\paragraph{Scale Invariant Feature Transform (SIFT)} We again downsample videos to 1 frame per second, then compute the Euclidean distance between the SIFT feature vectors of adjacent keyframes and select those with a difference greater than some threshold. For the segment-level summarization, we take the maximum, and for the whole-video summarization, we select keyframes whose differences from the previous keyframe are greater than the average.

\vspace{-2pt}
\paragraph{VSUMM} For VSUMM, we use a sampling rate the same as the fps of the video. 

\vspace{-2pt}
\paragraph{Keyframe Extraction} Using the video downsampled to 1 fps, we sampled one out of every 3 frames. We then use the differences of adjacent frames (represented as a CNN feature vector) to define scenes in the video. We set the threshold for drawing scene boundaries to 0.65. Using K-means and Euclidean distance, we cluster the keyframes per scene and then remove redundant candidate keyframes from the same scene using a threshold of 0.8.
\cite{jadon2019video}.

\vspace{-2pt}
\subsection{Text Summarization}

For textual summarization, we selected the following representative models as our baselines: BERT2BERT \cite{turc2019}, BART \cite{Lewis2020BARTDS} (BART-large-CNN and BART-large-XSUM), Distilbart \cite{Shleifer2020PretrainedSD}, T5 \cite{Raffel2019ExploringTL}, Pegasus  \cite{Zhang2019PEGASUSPW}, and Longformer Encoder-Decoder (LED) \cite{Beltagy2020LongformerTL}.

\vspace{-2pt}
\paragraph{BERT2BERT} 
Through an encoder-decoder architecture with the auto-regressive generation, we predict summaries from the extracted text at time $t$. Tokenized length $T_m$ and summary length $S_m$ are bounded as follows: $T_m \leq 512, 2 \leq S_m \leq 15$. Additional parameters include: \textit{truncation} = True, \textit{padding} = "max-length", \textit{skip\_special\_tokens} = True.
The pretrained model used can be found in the transformers library under BertTokenizerFast.

\vspace{-2pt}
\paragraph{BART-large-CNN} 
Using an encoder-encoder framework, the BART-large-CNN model first corrupts text with a noising function, then reconstructs this text with a CNN. Tokenized length $T_m$ and summary length $S_m$ are bounded as follows: $T_m \leq 512, 1 \leq S_m \leq 10$. Additional parameters include: \textit{num\_beams} = 2, \textit{clean\_up\_tokenization\_space} = True.
The pretrained Facebook model used can be found in the transformers library under BartforConditionalGeneration.

\vspace{-2pt}
\paragraph{BART-large-XSUM} 
Similar to BART-large-CNN, BART-large-XSUM employs a transformer-based neural machine translation architecture, effective in text generation and comprehension. Tokenized length $T_m$ and summary length $S_m$ are bounded as follows: $T_m \leq 512, 1 \leq S_m \leq 10$. Additional parameters include: \textit{num\_beams} = 2, \textit{skip\_special\_tokens} = True.

\vspace{-2pt}
\paragraph{Distilbart}
We use distilbart-cnn-6-6, which copies alternating layers from the BART-large-CNN model and integrates MSE loss from the tinybert model. Tokenized length $T_m$ and summary length $S_m$ are bounded as follows: $T_m \leq 512, 4 \leq S_m \leq 15$. A pretrained model from the transformers library was implemented: "ml6team/distilbart-tos-summarizer-tosdr".

\vspace{-2pt}
\paragraph{T5} 
T5 integrates supervised and unsupervised tasks in an encoder-decoder framework. We used the "t5-small" model, having optimized runtime compared to other T5 models. Summary length $S_m$ was bounded as follows: $2 \leq S_m \leq 15$. Additional parameters include: $num\_beams$ = 4, $no\_repeat\_ngram\_size$ = 2, $early\_stopping$ = True.

\vspace{-2pt}
\paragraph{Pegasus} 
Pegasus masks important sentences from the text, combining these into an output sequence to develop an informative summary; we used the ``pegasus-xsum" model, being the most fine-tuned. Summary length $S_m$ was bounded as follows: $2 \leq S_m \leq 15$. Additional parameters include: $padding$ = longest, $truncation$ = True.

\vspace{-2pt}
\paragraph{Longformer Encoder-Decoder (LED)} 
LED employs similar architectures to the BART model; however, it works better on longer input text (over 1024 tokens). We used the ``led-large-16384" model; some parameters include: $repetition\_penalty$ = 3.5, $encoder\_no\_repeat\_ngram\_size = $ 3, $early\_stopping$ = True, $no\_repeat\_ngram\_size$ = 3. Tokenizer length $S_m$ was bounded as follows: $16 \leq T_m \leq 256$.

\section{More Results and Discussions}\label{sec:appendix-more-results}

\begin{table}[tp]\small
\caption{Comparison of video temporal segmentation results.}
\vspace{-5pt}
\centering
\begin{adjustbox}{width=0.8\linewidth}
\begin{tabular}{lcccccc}
\toprule
Model  &Average Precision (AP) $\uparrow$ & F1 $\uparrow$ & $M_{iou}$ $\uparrow$ & Recall@3s $\uparrow$ & Recall@5s $\uparrow$ & Recall@10s  $\uparrow$ \\
\midrule
Histogram Intersect & 0.142 & 0.153 & 0.221 & 0.168 & 0.216 & 0.296 \\
Moment Invariant & 0.081 & 0.089 & 0.164 & 0.101 & 0.129 & 0.177  \\
Twin Comparison & 0.133 & 0.140 & 0.208 & 0.150 & 0.193 & 0.266 \\
PySceneDetect & 0.135 & 0.124 & 0.211 & 0.119 & 0.152 & 0.199 \\
LGSS & 0.243 & 0.352 & 0.216 & 0.163 & 0.216 & 0.272 \\
Ours & \textbf{0.503} & \textbf{0.423} & \textbf{0.223} & \textbf{0.325} & \textbf{0.341} & \textbf{0.366} \\
\bottomrule
\end{tabular}
\end{adjustbox}
\label{table:vts_results-app}
\vspace{-10pt}
\end{table}

\begin{table}[tp]\small
\caption{Comparison of video summarization results (whole video setting and segment-level setting).}
\vspace{-5pt}
\centering
\begin{adjustbox}{width=0.8\linewidth}
\begin{tabular}{llcccc|ccc}
\toprule
Setting &Model  & RMSE $\downarrow$  & PSNR $\uparrow$ &SSIM $\uparrow$    &SRE $\downarrow$   &Precision $\uparrow$ &Recall $\uparrow$ & F1 Score $\uparrow$ \\ 
\midrule
\multirow{5}{*}{Whole-video}
&Uniform & 0.479 & 4.044 & 0.076 & 49.808 & 0.077 & 0.100 & 0.049\\
&K-means  &0.348 &8.234 &0.055 &46.438 &0.072 &0.182 &0.103  \\   
&SIFT & 0.330 &8.497 & 0.046 & 45.949 & 0.047 & 0.125 & 0.059\\
&VSUMM &0.279 &9.226 &0.053 &44.862 &0.054 &0.259 &0.088 \\
&Ours&\textbf{0.112}&\textbf{25.280}&\textbf{0.697}&\textbf{23.550}&\textbf{0.320}&\textbf{0.290} & \textbf{0.321}\\
\midrule
\multirow{5}{*}{Segment-level}
&Uniform & 0.237 & 6.307 & 0.085 & 42.495 & 0.186 & 0.179 & 0.105\\
&K-means  &0.167 &10.123 &0.144 &46.533 &0.123 &0.172 &0.143   \\
&SIFT & 0.114 & 10.816 & 0.178 & 41.634 & 0.079 & 0.079 & 0.079\\
&VSUMM  & 0.122 & 18.818 & 0.258 & 41.601 & 0.160 & 0.207 & 0.171\\
&Ours &\textbf{0.091}&\textbf{36.370}&\textbf{0.698}&\textbf{23.430}&\textbf{0.333}&\textbf{0.275}&\textbf{0.255}\\
\bottomrule
\end{tabular}

\end{adjustbox}
\label{table:vs_results-app}
\vspace{-10pt}
\end{table}

\subsection{Results and Discussion}

\paragraph{Supervised training leads to more accurate video temporal segmentation results} 

The performance of video temporal segmentation has a great impact on the final performance, so in this section, we compare the performance of VTS with several baselines: 
Histogram Intersect \cite{Lee2005EvaluationOI}, 
Moment Invariant \cite{Huang2010AnalysisOH}, 
Twin Comparison \cite{Zhang1993AutomaticPO}, PySceneDetect \cite{PySceneDetect}, and LGSS \cite{Rao2020ALA}.  The results, displayed in Table~\ref{table:vts_results-app}, indicate that LGSS outperforms the other baselines but falls short when compared to our model. Both our method and LGSS are trained using a supervised approach, which leads to improved performance compared to unsupervised baselines. Moreover, our approach incorporates attention mechanisms, potentially contributing to better results.

\vspace{-5pt}
\paragraph{Supervised methods outperform unsupervised methods on video summarization}

In our video summarization study, we have chosen the following methods as our baseline comparisons: Uniform Sampling \cite{jadon2019video}, K-means Clustering \cite{Hartigan1979AKC}, Scale Invariant Feature Transform (SIFT) \cite{Morel2010IsT}, and VSUMM \cite{de2011vsumm}. The results, presented in Table~\ref{table:vs_results-app}, are under various evaluation metrics. For RMSE and SRE, lower values indicate better performance, whereas for the remaining metrics, higher values are desirable. From Table~\ref{table:vs_results-app}, we can observe that VSUMM showcases the strongest performance among the baseline methods, yet it still falls short compared to our proposed method. But we can conclude that supervised methods outperform unsupervised methods.

\begin{table}[tp]\small
\caption{Comparison of textual summarization results (whole video setting and segment-level setting).}
\vspace{-5pt}
\centering
\begin{adjustbox}{width=0.99\linewidth}
\begin{tabular}{llrrrrrrrr}
\toprule
Setting &Model  & BLEU-1 $\uparrow$ & ROUGE-1 $\uparrow$ & ROUGE-2 $\uparrow$  & ROUGE-L $\uparrow$ & METEOR $\uparrow$ & CIDEr $\uparrow$ & SPICE $\uparrow$ &BertScore $\uparrow$ \\
\midrule
\multirow{8}{*}{Whole-video}
&BERT2BERT \cite{turc2019} & 22.59  & 3.75 & 0.45 & 3.41 & 5.65 & 1.76 & 2.91 & 71.12 \\
&BART-large-CNN \cite{Lewis2020BARTDS} & 29.17  & 3.19 & 0.51 & 3.04 & 2.99 & 2.28 & 11.27 & 68.84\\
&BART-large-XSUM \cite{Lewis2020BARTDS} & 30.91  & 3.83 & 0.57 & 3.59 & 3.99 & 2.56 & 3.71 & 69.56 \\
&Distilbart \cite{Shleifer2020PretrainedSD}  & 26.46  & 3.87 & 3.87 & 0.47 & 3.59 & 2.25 & 4.16 & 69.37 \\
&T5 \cite{Raffel2019ExploringTL} & 25.39  & 3.51 & 0.43 & 3.21 & 4.51 & 1.97 & 5.66 & 70.38\\
&Pegasus \cite{Zhang2019PEGASUSPW} & 26.73  & 3.75 & 0.52 & 3.40 & 4.52 & 2.38 & 7.82 & 68.92 \\
&LED \cite{Beltagy2020LongformerTL} & 26.47  & 3.81 & 0.25 & 3.51 & 3.45 & 1.78 & 6.72 & 68.45\\
&Ours &  \textbf{32.61} & \textbf{9.41} & \textbf{2.86} & \textbf{9.15} & 4.01 & 4.01 & 10.11 & \textbf{74.46}\\
\midrule
\multirow{8}{*}{Segment-level}
&BERT2BERT \cite{turc2019} & 13.58  & 4.70 & 1.95 & 4.53 & 28.59 & 11.73 & 10.13 & 71.76 \\
&BART-large-CNN \cite{Lewis2020BARTDS} & 22.79  & 6.45 & 2.46 & 6.32 & 26.21 & 20.64 & 10.13 & 71.44\\ 
&BART-large-XSUM \cite{Lewis2020BARTDS} & 20.89  & 7.31 & 2.77 & 7.13 & 29.36 & 20.90 & 10.20 & 71.42 \\
&Distilbart \cite{Shleifer2020PretrainedSD} & 14.77  & 1.95 & 0.15 & 1.87 & 23.52 & 11.83 & 10.53 & 66.46\\
&T5 \cite{Raffel2019ExploringTL} & 16.48  & 6.17 & 3.03 & 5.99 & 28.22 & 20.96 & 10.35 & 71.95 \\
&Pegasus \cite{Zhang2019PEGASUSPW} & 16.17  & 3.41 & 0.96 & 3.29 & 29.82 & 17.26 & 10.39 & 67.81\\
&LED \cite{Beltagy2020LongformerTL} & 16.03  & 3.80 & 0.60 & 3.64 & 29.81 & 15.85 & 10.99 & 68.46\\
&Ours &  \textbf{23.36} & \textbf{13.61} & \textbf{4.58} & \textbf{13.24} & \textbf{30.01} & \textbf{21.06} & 10.28 & \textbf{85.19}\\ 
\bottomrule
\end{tabular}

\label{table:ts_results-app}
\end{adjustbox}
\vspace{-10pt}
\end{table}

\begin{table}[tp]\small
\caption{Comparison of MSMO results.}
\vspace{-5pt}
\centering
\begin{adjustbox}{width=0.99\linewidth}
\begin{tabular}{l|cccccccccccc} 
\toprule
\multirow{2}{*}{Methods}   & \multicolumn{5}{c}{Text} &\multicolumn{5}{c}{Video}   \\ 
\cmidrule(r){2-6}\cmidrule(r){7-11} 
& BLEU $\uparrow$ & METEOR $\uparrow$ & CIDEr $\uparrow$ & SPICE $\uparrow$ &BertScore $\uparrow$ &PSNR $\uparrow$ &SSIM $\uparrow$ &Precision  $\uparrow$ &Recall $\uparrow$ &F1 Score $\uparrow$ \\ 
\midrule
LGSS + VSUMM + T5 & 27.35 & 24.32 & 3.94 & 5.57 & 62.77 &16.234  &0.198  &0.143  &0.152  &0.147   \\ 
LGSS +  VSUMM + BART-large-XSUM & 24.83 & 24.12 & 4.37 & 8.86  &39.20 &16.234  &0.198  &0.143  &0.152  &0.147    \\
LGSS +  VSUMM + BERT2BERT & 13.26 & 34.83 & 3.68 & 9.23& 64.34  &16.234  &0.198  &0.143  &0.152  &0.147   \\
LGSS +  VSUMM + BART-large-CNN & 24.93 & 32.61 & 4.18 & 11.84 & 64.44 &16.234  &0.198  &0.143  &0.152  &0.147 \\
Ours& \textbf{33.36} & 30.31 & 4.06 & 10.28 & \textbf{85.19} & \textbf{36.370} & \textbf{0.298} & 0.133 & \textbf{0.275} &  \textbf{0.155} \\
\bottomrule
\end{tabular}
\label{table:msmo_results-app}
\end{adjustbox}
\vspace{-10pt}
\end{table}

\vspace{-5pt}
\paragraph{Pretrained large language models can still do well in text summarization}

In the context of textual summarization, we have considered a set of representative models as our baseline comparisons: BERT2BERT \cite{turc2019}, BART \cite{Lewis2020BARTDS} (including BART-large-CNN and BART-large-XSUM), Distilbart \cite{Shleifer2020PretrainedSD}, T5 \cite{Raffel2019ExploringTL}, Pegasus \cite{Zhang2019PEGASUSPW}, and Longformer Encoder-Decoder (LED) \cite{Beltagy2020LongformerTL}. The performance of these models is summarized in Table~\ref{table:ts_results-app}. Among the baselines, T5, BART-large-XSUM, BART-large-CNN, and BERT2BERT exhibit superior performance, with T5 demonstrating relatively better results across various text evaluation metrics. It is worth noting that the ROUGE score may not effectively capture performance differences compared to other evaluation metrics, which tend to provide more meaningful variations in performance.

\vspace{-5pt}
\paragraph{MSMO results may depend on segmentation results and summarization methods}

In the field of MSMO, we encountered limitations in accessing the codebases of existing works such as \cite{Zhu2018MSMOMS,hori2019end,zadeh-etal-2017-tensor,chen2018abstractive,Fu2021MMAVSAF,Fu2020MultimodalSF}. Therefore, we independently implemented several baselines to evaluate their performance on the MMSum dataset. For this purpose, we utilized LGSS as the segmentation backbone, VSUMM as the video summarizer, and selected text summarizers that exhibited the best performance in text summarization. The results are presented in Table~\ref{table:msmo_results-app}. Based on the findings, it is evident that the aforementioned combination approaches still fall short in comparison to our proposed method. This also indicates that the accuracy of temporal segmentation is crucial prior to generating summaries, highlighting it  as a critical step and task preceding MSMO.

\section{More Related Work}\label{sec:appendix-related-work}
\vspace{-5pt}
\paragraph{Video Temporal Segmentation} aims at splitting the video into segments based on predefined rules,  which is a fundamental step in video analysis. Previous work either formed a classification problem to detect the segment boundaries in a supervised manner \cite{Sidiropoulos2011TemporalVS,Zhou2013HierarchicalAC,Poleg2014TemporalSO,Sokeh2018SuperframesAT,Aakur2019APP} or in the unsupervised way \cite{Gygli2014CreatingSF,Song2015TVSumSW}. Temporal segmentation of actions in videos is also widely explored in previous works \cite{Wang2019TemporalSN,Zhao2017TemporalAD,Lea2017TemporalCN,Kuehne2020AHR,Sarfraz2021TemporallyWeightedHC,Wang2020BoundaryAwareCN}. Video shot boundary detection and scene detection tasks are also relevant and has been explored in many previous studies \cite{Hassanien2017LargescaleFA,Hato2019FastAF,Rao2020ALA,Chen2021ShotCS,Zhang2021BetterLS}, which aim at finding the visual change or scene boundaries. 

\vspace{-5pt}
\paragraph{Video Summarization} aims at extracting key moments that summarize the video content by selecting the most informative and vital parts, which lie in two directions: unimodal and multimodal approaches. Unimodal methods only use the visual modality, while multimodal methods exploit the available textual metadata and learn semantic or category-driven summarization. The summary usually contains a set of representative video keyframes that have been stitched in chronological order \cite{Apostolidis2021VideoSU}. Traditional video summarization methods only use visual information, while recently, some category-driven or supervised approaches were proposed to generate video summaries with video-level labels \cite{song2015tvsum,zhou2018deep,xiao2020convolutional,Zhou2018VideoSB,Haopeng2022ProgressiveVS,Narasimhan2022TLDWSI}. 

\vspace{-5pt}
\paragraph{Text Summarization} takes textual metadata, i.e., documents, articles, tweets, etc, as input, and generates textual summaries in two directions: abstractive or extractive summarization. Abstractive methods select words based on semantic understanding, even the words may not appear in the source \cite{tan2017abstractive,Abigail2017}. Extractive methods attempt to summarize language by selecting a subset of words that retain the most critical points, which weights the essential part of sentences to form the summary \cite{narayan2018ranking, wu2018learning}.  
Recently, the fine-tuning approaches have improved the quality of generated summaries based on pre-trained language models in a wide range of tasks \cite{liu-lapata-2019-text, zhang-etal-2019-hibert}.

\section{More Thumbnail Results}\label{sec:appendix-thumbnail-results}

\begin{figure}[htp]
  \centering
  \includegraphics[width=0.7\linewidth]{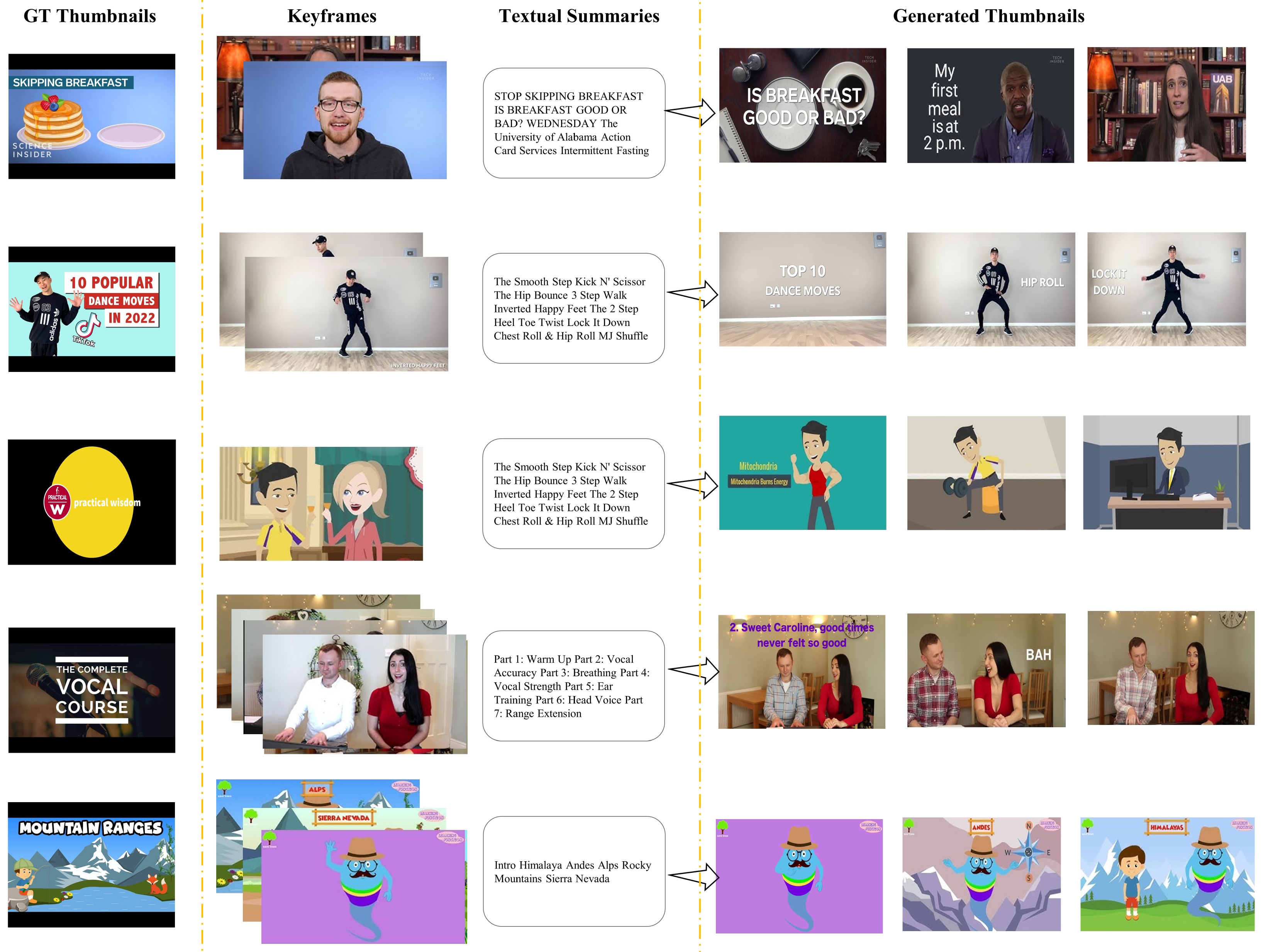}
  \vspace{-5pt}
  \caption{More comparison of GT thumbnails and our generated ones [3/3].}
  \label{fig:example3}
  \vspace{-10pt}
\end{figure}

\begin{figure}[htp]
  \centering
  \includegraphics[width=0.7\linewidth]{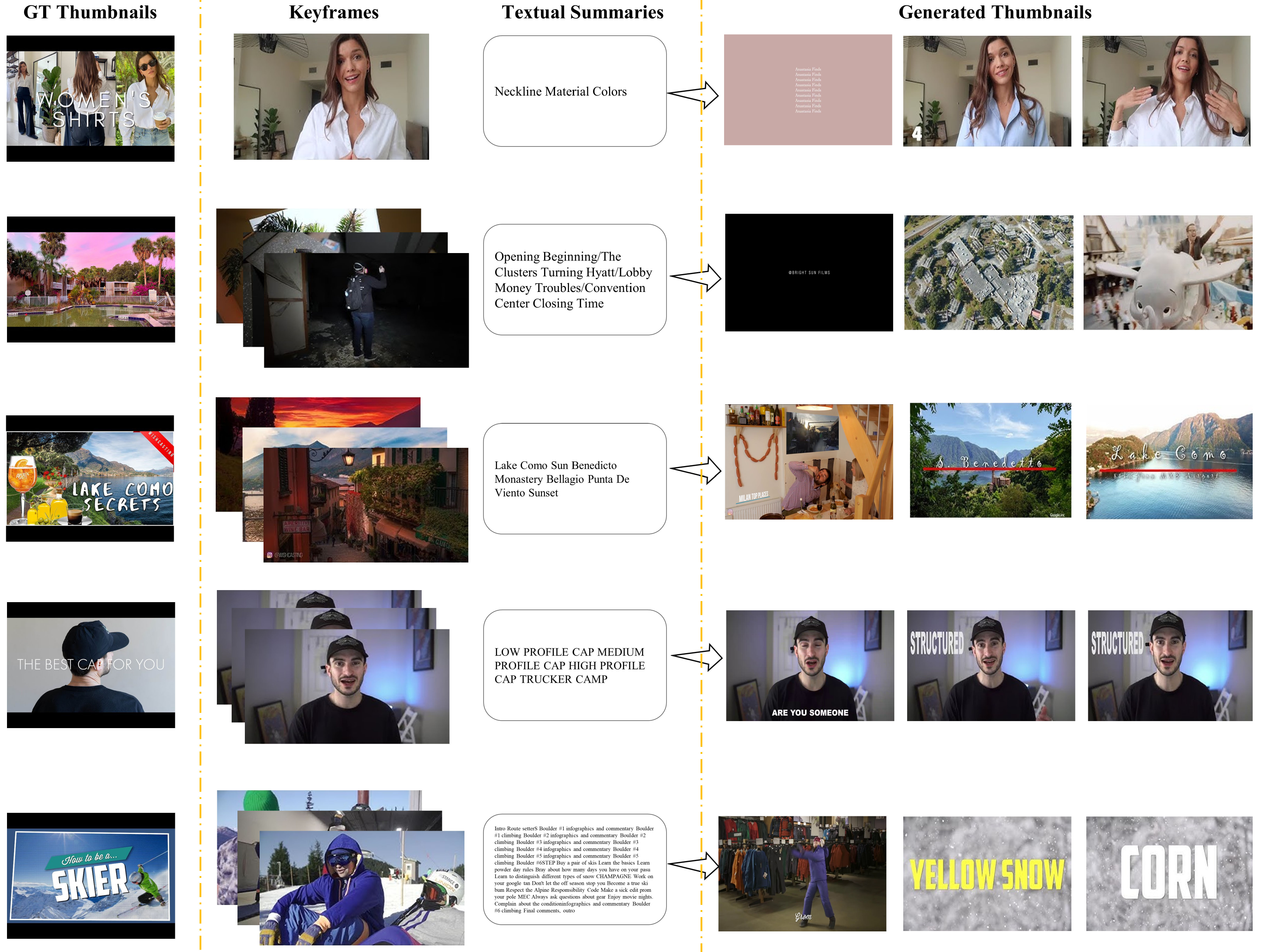}
  \vspace{-5pt}
  \caption{More comparison of GT thumbnails and some ``bad" generated ones.}
  \label{fig:example4}
  \vspace{-10pt}
\end{figure}

In the following Figures~\ref{fig:example1}, \ref{fig:example2}, and \ref{fig:example3}, we show more comparisons of our generated thumbnails with Ground-Truth (GT) thumbnails provided by the authors of the video. We can find that our generated thumbnails can be very informative. Besides, we also provided some not-good examples in Figures~\ref{fig:example4}, showing the potential of this new task and lots of room for improvement. 

\clearpage

\begin{figure}[htp]
  \centering
  \includegraphics[width=0.7\linewidth]{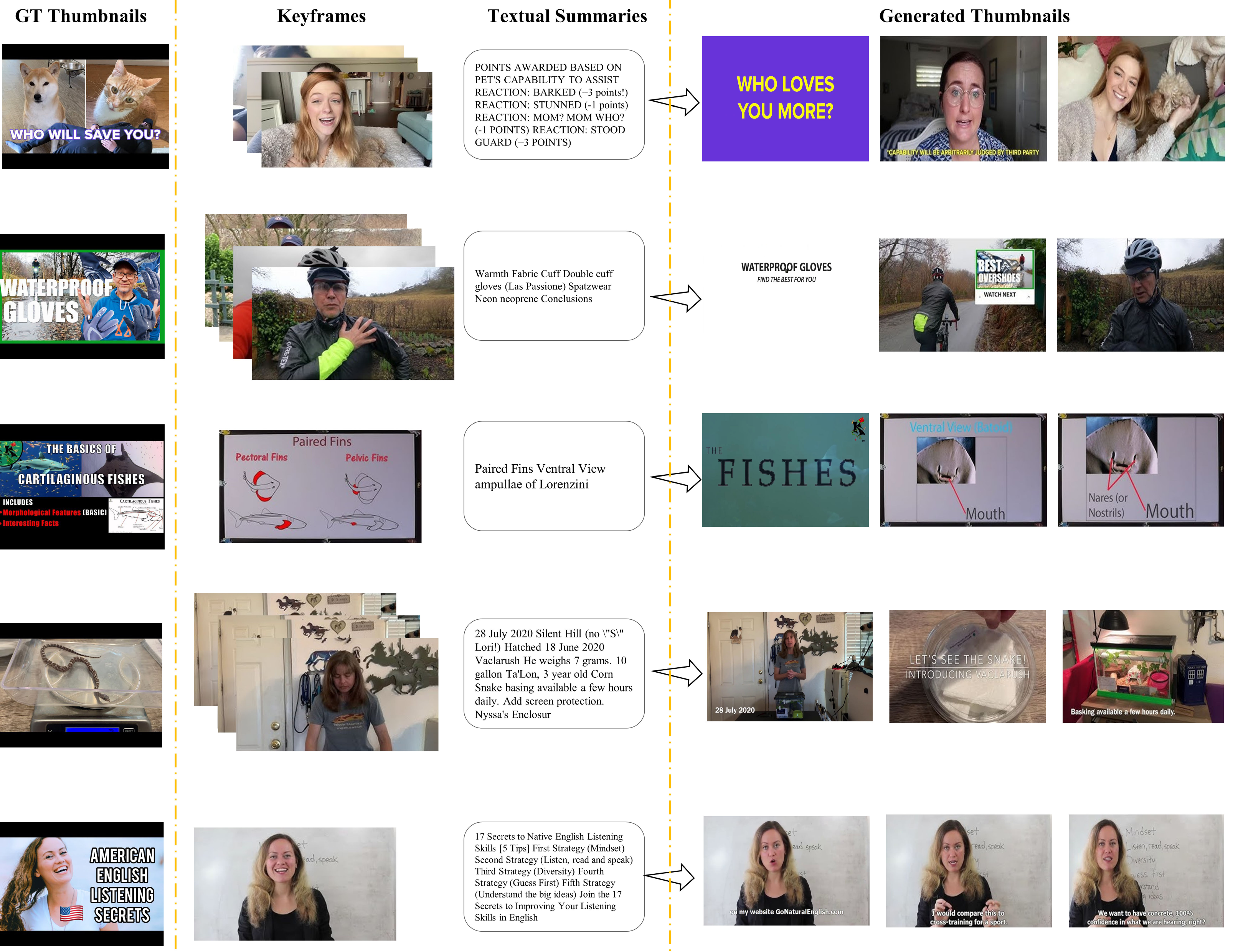}
  \vspace{-5pt}
  \caption{More comparison of GT thumbnails and our generated ones [1/3].}
  \label{fig:example1}
  \vspace{-10pt}
\end{figure}

\begin{figure}[htp]
  \centering
  \includegraphics[width=0.7\linewidth]{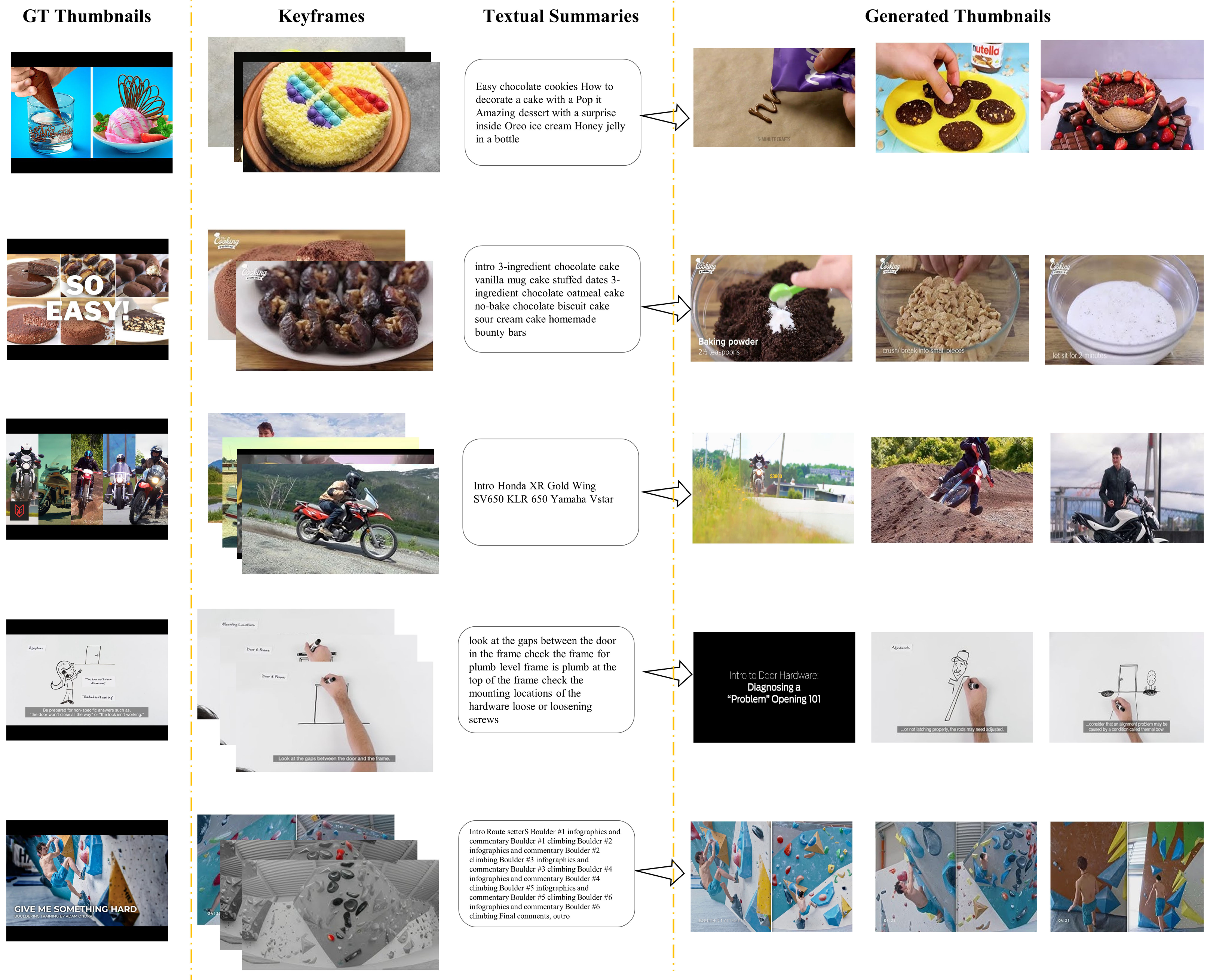}
  \vspace{-5pt}
  \caption{More comparison of GT thumbnails and our generated ones [2/3].}
  \label{fig:example2}
  \vspace{-10pt}
\end{figure}

\end{document}